Review Article

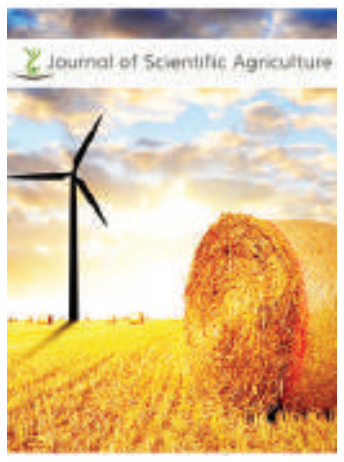

# Leveraging deep learning for plant disease identification: a bibliometric analysis in SCOPUS from 2018 to 2024

**Enow Takang Achuo Albert*, Ngalle Hermine Bille, Ngonkeu Mangaptche Eddy Leonard**

Department of Plant Biology, Faculty of Science, University of Yaoundé I, P.O. Box 812, Yaoundé, Center Region, Cameroon

***Corresponding Author:**
Enow Takang Achuo Albert
E-mail: albert.enow@facsciences-uy1.cm

## ABSTRACT

This work aimed to present a bibliometric analysis of deep learning research for plant disease identification, with a special focus on generative modeling. A thorough analysis of SCOPUS-sourced bibliometric data from 253 documents was performed. Key performance metrics such as accuracy, precision, recall, and F1-score were analyzed for generative modeling. The findings highlighted significant contributions from some authors Too and Arnal Barbedo, whose works had notable citation counts, suggesting their influence on the academic community. Co-authorship networks revealed strong collaborative clusters, while keyword analysis identified emerging research gaps. This study highlights the role of collaboration and citation metrics in shaping research directions and enhancing the impact of scholarly work in applications of deep learning to plant disease identification. Future research should explore the methodologies of highly cited studies to inform best practices and policy-making.

**KEYWORDS:** Deep learning, Plant disease diagnosis, Generative modeling, Bibliometric analysis

## INTRODUCTION

Deep learning has emerged as a transformative technology in agricultural science, particularly for the identification of plant diseases. This approach leverages advanced algorithms, primarily Convolutional Neural Networks (CNNs), to analyze images of plants and accurately diagnose diseases that threaten crop health and yield (Mohanty *et al.*, 2016; Guo *et al.*, 2020; Saleem *et al.*, 2020; Ahmed & Yadav, 2023; Jung *et al.*, 2023; Shoaib *et al.*, 2023; Pacal *et al.*, 2024). Plant diseases pose a significant threat to global food security, leading to substantial yield losses and economic impacts on agriculture. Traditional methods of disease identification often rely on visual assessments by trained professionals, which can be time-consuming, subjective, and prone to errors (Jafar *et al.*, 2024). As a result, there is a pressing need for automated systems that can provide rapid and accurate disease detection to support farmers and agricultural experts in managing crop health effectively.

Deep learning models, especially CNNs, have been shown to outperform traditional methods in terms of accuracy and efficiency. These models can learn hierarchical representations from raw image data, enabling them to identify complex patterns associated with various plant diseases. Recent studies have demonstrated that CNNs can achieve accuracy rates as high as 99.35% when classifying images of diseased and healthy plants. The architecture of CNNs typically includes layers for feature extraction and classification, allowing them to process visual information effectively. Other deep learning models like Generative Adversarial Networks (GANs) are also utilized for enhancing dataset diversity by generating synthetic images of diseased plants, improving model robustness (Saleem *et al.*, 2019; Hasan *et al.*, 2020; Nigam & Jain, 2020; Bhuvana & Mirnalinee, 2021; Chen *et al.*, 2022; Ahmad *et al.*, 2023; Benfenati *et al.*, 2023; Dhaka *et al.*, 2023; Omer *et al.*, 2023; Ramanjot *et al.*, 2023; Shoaib *et al.*, 2023; Ghafar *et al.*, 2024; Ngugi *et al.*, 2024; Rodríguez-Lira *et al.*, 2024; Saleem *et al.*, 2024; Yang *et al.*, 2024; Yuan *et al.*, 2024).

Multi-task learning approaches allow simultaneous learning of multiple related tasks, such as identifying host species and associated diseases, thereby enhancing the model's ability to generalize across different contexts (Gu *et al.*, 2022; Dai *et al.*, 2023; Song *et al.*, 2023; Zhao *et al.*, 2023; Hemalatha &

Jayachandran, 2024; Liu *et al.*, 2024). The effectiveness of deep learning models in plant disease identification significantly depends on the availability of large, high-quality datasets. Notable datasets include PlantVillage (Hughes & Salathe, 2016), which contains thousands of images across various crop species and diseases, and the Rice Leaf Disease Dataset (Prajapati *et al.*, 2017), which focuses specifically on rice crops. Custom datasets introduced in recent studies cover a broader range of host species and diseases, facilitating improved model training and validation. However, several challenges remain in this field. High-quality labeled datasets are crucial but often limited in scope or accessibility. Additionally, variability in image quality can affect model performance; thus, standardized imaging protocols are necessary. Understanding how models make decisions is essential for gaining trust among users; techniques like eXplainable Artificial Intelligence (XAI) are being explored to address this need (Thakur *et al.*, 2022; Sagar *et al.*, 2023; Amara *et al.*, 2024; Jafar *et al.*, 2024; Natarajan *et al.*, 2024).

Several arguments have been used to support the use of deep learning in plant disease identification. (1) Enhanced accuracy and efficiency: Deep learning models, particularly Convolutional Neural Networks (CNNs), have shown remarkable accuracy in identifying plant diseases by automatically learning features from images. This capability significantly reduces the manual effort required for disease detection, which is often time-consuming and prone to human error. (2) Early detection and intervention: The ability of deep learning systems to process large datasets enables early identification of diseases, allowing for timely intervention. Early detection is critical in agriculture as it can prevent the spread of diseases, thereby safeguarding crop yields and quality (Saleem *et al.*, 2019; Nguyen *et al.*, 2021; Andrew *et al.*, 2022; Alzahrani & Alsaade, 2023; Mustofa *et al.*, 2023; Rehana *et al.*, 2023; González-Rodríguez *et al.*, 2024; Radočaj *et al.*, 2024). (3) Scalability and automation: Deep learning technologies facilitate the automation of plant disease identification processes, making them scalable. This means that farmers can monitor large areas of crops efficiently without the need for extensive labor, thus optimizing resource use and reducing costs. (4) Integration with smart farming: The integration of deep learning with smart farming technologies enhances decision-making processes in agriculture. By providing data-driven insights, these systems support advanced analyses and planning, which are essential for modern agricultural practices (Altalak *et al.*, 2022; Dhaka *et al.*, 2023; Vankara *et al.*, 2023; Ali *et al.*, 2024; Krishna *et al.*, 2024; Lebrini & Ayerdi Gotor, 2024; Nagaraj *et al.*, 2024; Nanavaty *et al.*, 2024; Sajitha *et al.*, 2024). (5) Adaptability to diverse conditions: Deep learning models can be trained on various datasets, allowing them to adapt to different environmental conditions and plant species. This adaptability ensures that the models remain effective across diverse agricultural settings, contributing to global food security efforts.

Conducting a bibliometric analysis in the domain of deep learning for plant disease identification is crucial for more than a few reasons, as it provides comprehensive insights into the evolution, trends, and impact of research in this rapidly advancing field. (1) Understanding research trends: Bibliometric analysis allows researchers to identify and quantify the growth of publications related to deep learning applications in plant disease detection. By analyzing publication trends over time, researchers can gauge the increasing interest in this area, which is vital given the growing need for sustainable agricultural practices and food security. (2) Identifying key contributors and collaborations: A bibliometric analysis also highlights influential authors, institutions, and countries leading the research efforts in deep learning for plant disease identification. By mapping out co-authorship networks and institutional collaborations, researchers can identify potential partners for future studies or projects. Such collaboration is essential as it often leads to enhanced research quality and innovation through shared expertise and resources. (3) Keyword analysis and research gaps: Another critical aspect of bibliometric analysis is keyword analysis, which helps in identifying trending topics within the field. By examining frequently used keywords, researchers can uncover emerging areas of interest or gaps that require further exploration. This insight is particularly beneficial for guiding future research directions and ensuring that new studies address underexplored aspects of deep learning applications in plant disease identification. (4) Evaluating impact and citation analysis: Bibliometric analysis also facilitates citation analysis, which assesses the impact of specific studies or authors within the academic community. Understanding which publications are most cited can help researchers identify foundational works that have shaped the field or innovative studies that have introduced novel methodologies. This evaluation can inform scholars about best practices and methodologies that have proven effective in previous research. (5) Facilitating policy making and funding decisions: Finally, bibliometric analysis serves as a valuable tool for policymakers and funding agencies by providing evidence-based insights into research trends and priorities. By understanding where significant advancements are being made in deep learning for plant disease identification, stakeholders can allocate resources more effectively to areas that promise high returns on investment in terms of agricultural productivity and sustainability.

This work was therefore carried out mindful of the afore-attributed benefits to bibliometric analysis, with an additional focus on reviewing studies which have employed generative modeling technology in the domain of plant disease identification.

## MATERIALS AND METHODS

### Bibliometric Data Acquisition

Bibliometric data was acquired from SCOPUS using the following five search filters: (1) The query - "deep AND learning AND for AND plant AND disease AND identification OR classification OR recognition AND generative AND adversarial AND network AND for AND plant AND disease AND identification AND variational AND autoencoders AND for AND plant AND disease AND identification". (2) The period - 2018 to 2024. (3) The document type - article, conference paper, book chapter. (4) The language - English. (5)

The source type - journal, conference proceeding, book, book series. Following the search, 253 documents were returned and their bibliometric data were downloaded in CSV format.

## Software Used

VOSviewer 1.6.20 was used for bibliometric analyses. Python 3.11.8 was used for computing and visualizing publication and citation trends.

## Co-Authorship Analysis

Co-authorship analysis is a bibliometric method that examines the collaborative relationships between authors in academic research, revealing how they work together on publications and highlighting the social networks formed through collaboration. This type of analysis is important because it helps map out research collaboration patterns, illustrating the collective nature of knowledge production in modern science. Additionally, it assesses how collaboration impacts research productivity and citation rates, as studies indicate that co-authored works often achieve higher output and impact. Co-authorship networks can also identify emerging trends and influential areas of research, providing valuable insights for funding agencies, policymakers, and academic institutions on where to direct resources and support.

Full counting was specified. With respect to thresholds, (1) the minimum number of documents of an author was set to 3. (2) the minimum number of citations of an author was set to zero. There was a total of 900 authors, of which 23 met the thresholds and only 14 were connected to each other. When the minimum number of documents of an author was set to 1 and all other parameters unaltered, all 900 authors met the threshold, and 164 authors were connected to each other.

## Keyword Co-occurrence Analysis

Keyword co-occurrence analysis is a bibliometric method that examines the frequency with which specific keywords appear together in a set of publications. This technique helps identify key themes, trends, and the relationships between different research topics within the literature. Its importance lies in its ability to reveal underlying patterns in research areas, facilitating a better understanding of how concepts are interconnected and guiding future research directions. By analyzing co-occurring keywords, researchers can efficiently map out the landscape of a field, highlighting emerging topics and influential areas of study.

Full counting was also specified. With respect to thresholds, the minimum number of occurrences of a keyword was set at 5. Of the 1524 keywords identified, only 110 met the threshold.

## Author-based Citation Analysis

Author-based citation analysis focuses on measuring the impact and influence of individual authors by counting the frequency with which their works are cited in other scholarly publications. This method not only identifies influential authors within a specific field but also helps trace the intellectual connections and trends in research over time. Its importance lies in providing insights into the scholarly impact of authors, guiding research funding, promotion decisions, and understanding the evolution of scientific disciplines through the relationships formed by citations.

With respect to thresholds, (1) the minimum number of documents of an author was set to 1. (2) the minimum number of citations of an author was set to 10. Of the 900 authors, 340 met the thresholds.

## Country-based Citation Analysis

Country-based citation analysis is a bibliometric method that evaluates the citation patterns of scholarly works attributed to specific countries. This approach helps to assess the scientific output, impact, and collaboration trends of nations within the global research landscape. Its importance lies in providing insights into national research strengths, facilitating comparisons between countries, and informing policy decisions regarding funding and resource allocation in science and technology.

With respect to thresholds, (1) the minimum number of documents of a country was set to 1. (2) the minimum number of citations of a country was set to 1. Of the 40 countries, 38 met the thresholds. However, only 9 countries were found to collaborate with at least another country.

## Document-based Bibliographic Coupling Analysis

Document-based bibliographic coupling analysis is a bibliometric method that identifies relationships between documents based on their shared citations. Specifically, it occurs when two documents cite one or more common sources, indicating a potential similarity in subject matter. This analysis is crucial in bibliometric studies as it helps researchers uncover connections among scholarly works, facilitating the exploration of related literature and the identification of research trends over time. By measuring the strength of these couplings, researchers can effectively cluster documents and gain insights into the intellectual structure of a field.

Full counting was employed. With respect to thresholds, the minimum number of citations of a document was set to 10. Of the 253 documents, only 85 met the threshold. Also, 84 out of the 85 documents were coupled.

## Author-based Co-citation Analysis

Author-based co-citation analysis (ACA) is a bibliometric method that examines the frequency with which two authors are cited together in scholarly literature. This analysis helps to reveal the intellectual structure of a particular academic field by mapping relationships between authors based on their co-citation patterns. The significance of ACA lies in its ability to identify clusters of related research, track the evolution of scientific disciplines, and highlight influential authors, thereby facilitating a deeper understanding of knowledge development and dissemination within various domains.

Full counting was specified. With respect to thresholds, the minimum number of citations of an author was set to 20. Of the 12564 authors cited, only 128 met the threshold.

## Publication and Citation Trends

Publication and citation trends in bibliometric analysis refer to the statistical examination of patterns and changes in academic publishing over time, focusing on aspects such as the volume of publications, authorship, citation rates, and thematic evolution within specific fields. This analysis is crucial as it helps researchers and institutions understand the dynamics of scholarly communication, identify emerging research areas, and assess the impact of published works. These were computed and visualized with Python 3.11.8.

## Generative Modeling for Plant Disease Identification

Generative models are important tools in image-based modeling for several reasons: (1) Data Augmentation: Generative models can create synthetic images that augment existing datasets. This is especially beneficial in scenarios where acquiring labeled data is difficult or expensive. By generating additional training samples, these models help improve the robustness and accuracy of classification algorithms, enabling better performance in identifying diseases from images (Antoniou *et al.*, 2018; Kebaili *et al.*, 2023; Wang *et al.*, 2023a; Zheng *et al.*, 2023; Alimisis *et al.*, 2024; Che *et al.*, 2024; Chen *et al.*, 2024; Fu *et al.*, 2024; Lingenberg *et al.*, 2024; Rahat *et al.*, 2024). (2) Addressing Class Imbalance: Many image-based datasets suffer from class imbalance, where certain diseases are underrepresented. Generative models can produce synthetic images for underrepresented classes, thus balancing the dataset. This balance is essential for training deep learning models effectively, as it reduces bias towards more prevalent classes and enhances overall classification accuracy (Huang & Jafari, 2021; Mirza *et al.*, 2021; Pan *et al.*, 2024; Wan *et al.*, 2024). (3) Cost-Effectiveness: Using generative models to create synthetic datasets can be more cost-effective than manual data collection and labeling. This approach reduces the time and resources needed for gathering large datasets while still allowing for comprehensive training of machine learning algorithms. This efficiency is particularly valuable in agricultural research where funding and resources may be limited.

With respect to plant diseases, two types of generative models stand out – Auto-Encoders and Generative Adversarial Networks.

### *Auto-encoders*

Autoencoders are a class of artificial neural networks designed for unsupervised learning tasks, particularly suited for generating compressed representations of data, such as images (Makhzani *et al.*, 2016; Bank *et al.*, 2021; Bourlard & Kabil, 2022; Gulamali *et al.*, 2022; Michelucci, 2022; Cunningham *et al.*, 2023; Lee, 2023; Martino *et al.*, 2023; Bunker *et al.*, 2024; Gao *et al.*, 2024). They consist of three main components: the encoder, the code (or latent representation), and the decoder. The encoder compresses the input data into a lower-dimensional representation known as the latent space or code. It maps the input X to a code Z through a series of transformations, progressively reducing dimensionality across multiple layers. For instance, in a typical setup for image data, the input layer represents pixel values (e.g., 28x28 pixels for MNIST), followed by hidden layers with decreasing numbers of neurons, culminating in a bottleneck layer that forms the smallest representation of the latent space. The decoder mirrors this architecture, starting from the bottleneck layer and gradually increasing dimensionality back to that of the original input. Each layer in the decoder corresponds to a layer in the encoder, ensuring symmetry (Zhao *et al.*, 2016; Rudolph *et al.*, 2019; D'Angelo & Palmieri, 2021; Baig *et al.*, 2023; Chen & Guo, 2023; Chiba *et al.*, 2023; Merkelbach *et al.*, 2023; Moyes *et al.*, 2023; Bertrand *et al.*, 2024; Hu *et al.*, 2024). The performance of an autoencoder is evaluated using a loss function that measures how well the output approximates the original input, with common choices including mean squared error (MSE) for continuous data and binary cross-entropy for binary or normalized data. The objective is to minimize this loss during training through backpropagation, adjusting weights in both the encoder and decoder (Creswell *et al.*, 2017; Abrar & Samad, 2022; Berahmand *et al.*, 2024).

Different types of autoencoders have been developed to enhance their capabilities, two of which shall be discussed in this review - Variational Auto-Encoders (VAEs) and Denoising Convolutional Variational Auto-Encoders DC-VAEs.

VAEs are a class of generative models that merge concepts from deep learning and Bayesian inference. Introduced by Diederik P. Kingma and Max Welling in their influential 2013 paper "Auto-Encoding Variational Bayes" (Kingma & Welling, 2022), VAEs aim to learn the underlying probability distribution of data, enabling the generation of new samples that resemble the training data. The architecture of VAEs consists of two primary components: an encoder and a decoder. The encoder maps input data to a latent space, producing parameters (mean and variance) of a probability distribution instead of fixed points. This probabilistic representation allows the model to capture latent variables, typically assuming a Gaussian distribution. The decoder, on the other hand, takes samples from this latent space and reconstructs the original input data, facilitating the generation of new data points during inference. A distinctive feature of VAEs is their latent space representation. Unlike traditional autoencoders that map inputs to discrete points in latent space, VAEs ensure that each input corresponds to a distribution, which allows for smooth interpolations between data points and the generation of new samples by sampling from the learned distributions (Kipf & Welling, 2016; Casale *et al.*, 2018; Kingma & Welling, 2019, 2022; Cemgil *et al.*, 2020; Ghosh *et al.*, 2020; Skopek *et al.*, 2020; Doersch, 2021; Sandfort *et al.*, 2021; Singh & Ogunfunmi, 2021; Bandyopadhyay *et al.*, 2022; Girin *et al.*, 2022; Gomari *et al.*, 2022; Manduchi *et al.*, 2023; Papadopoulos & Karalis, 2023).

DC-VAEs represent an advanced extension of the traditional Variational Autoencoder (VAE) framework, integrating principles of denoising and convolutional neural networks to enhance the model's robustness and performance in generating high-quality data. The primary goal of DC-VAEs is to effectively reconstruct data that may have been corrupted by noise, while simultaneously learning a useful latent representation of the input data. The architecture of a DC-VAE consists of two main components: the encoder and the decoder. The encoder transforms the input data into a latent space representation, while the decoder reconstructs the original data from this latent representation. In contrast to standard VAEs, which typically use fully connected layers, DC-VAEs leverage convolutional layers that are particularly effective for processing image data due to their ability to capture spatial hierarchies and local patterns. This convolutional approach allows DC-VAEs to learn more complex features from images, making them suitable for tasks such as image denoising, generation, and anomaly detection. A crucial innovation in DC-VAEs is the incorporation of a denoising mechanism. During training, instead of using clean input data directly, the model is trained on noisy versions of the data. This process involves adding noise to the input images and then training the encoder to recover the original images from these corrupted inputs. By doing so, DC-VAEs not only learn to generate new samples but also become adept at removing noise from existing samples. This dual capability enhances their performance in real-world applications where data is often imperfect or noisy. Mathematically, DC-VAEs utilize a loss function that combines reconstruction loss and a regularization term derived from Kullback-Leibler (KL) divergence, similar to traditional VAEs. The reconstruction loss measures how well the decoder can reconstruct the original data from its latent representation, while the KL divergence encourages the learned latent distribution to resemble a prior distribution (usually Gaussian). This balance between reconstruction accuracy and latent space regularization is crucial for ensuring that the model generalizes well to unseen data. The effectiveness of DC-VAEs has been demonstrated across various applications, particularly in image processing tasks such as generating high-fidelity images from noise or incomplete data. They have also shown promise in fields like medical imaging, where denoising capabilities can significantly improve diagnostic accuracy by enhancing image quality without losing critical information (Lee *et al.*, 2018; Chen & Shi, 2019; Cakmak *et al.*, 2020; Lei & Yang, 2021; Prakash *et al.*, 2021; Venkataraman, 2022; Chandrakala & Vishnika, 2024; Cheng *et al.*, 2024; Giuliano *et al.*, 2024; Iakovenko & Bondarenko, 2024; Qin *et al.*, 2024).

***Generative adversarial networks***

A Generative Adversarial Network (GAN) (Goodfellow *et al.*, 2014) is not a single model. It is a combination of two models. The first model is a Generator model (*G*), while the second is a Discriminator model (*D*). *D* learns the conditional probability of the target variable given the input variable, expressed as:

$$P(Y \mid X = x)$$

Most common examples are logistic regression, linear regression, etc. *G* learns the joint probability distribution of the input variable and the output variable, expressed as:

$$\begin{aligned} & P(X,Y) \\ & = P(X \mid Y)P(Y) \\ & = P(Y \mid X)P(X) \end{aligned}$$

If the model wants to make a prediction, then it uses bayes theorem and computes the conditional probability of the target variable, given the input variable, expressed as:

$$P(Y \mid X) = \frac{P(X \mid Y)}{P(X)}$$

The most common example is the Naïve Bayes model. The biggest advantage of generative models (*G*) over discriminatory models (*D*) is that we can use generative models to make new instances of data, since in generative models, we are learning the distribution function of the data itself. This is not possible with a discriminator. We use the generator to produce 'fake' data points, then use the discriminator to determine if a given data point is an original data point, or it has been produced by a generator. The *G* and *D* models work in an adversarial setup, i.e., they compete with each other, then get better and better at their jobs.

It is important to examine the high-level structure of a GAN. *G* and *D* are nothing but multi-layer neural networks (MLNNs). MLNNs are used here because they can approximate any function. This can be proven from the *universal approximation theorem*. Let the weights of G and D be and $\theta_d$, respectively. Suppose that the distribution function of the original data is:

$$\rho_{data}(x)$$

In reality, it is not really possible to draw or even mathematically compute the distribution function of the original data. This is due to the fact that we input data types like voice, images and videos, which are very high-dimensional. Let us consider the normal distribution as a noise distribution:

$$\rho_z(z)$$

Consider that we randomly sample data from the noise distribution and feed it to G. G will then output $G(z)$. The distribution of $G(z)$ is described as the same distribution with the original data:

$$\rho_g(x)$$

This is described as such, since the domain of the original data is the same as the range of . $G(z)$ This is important, since the goal is to try to replicate the original data. It is important to retain the following expressions as the distributions of the original data, the noise and the output of the generator function, respectively:

$$\rho_{data}, \rho_z, \rho_g$$

Consider that the labeled reconstructed data and the original data are passed to the discriminator, *D*. *D* would learn to return a single output, which would inform about the probability of the input belonging to the original data. Then, when next an unlabeled input is presented to *D*, it would try to determine if it is from the generator or from the original data class. As *D* improves, *G* would also be trained to better learn how to deceive *D*. So, the objectives of *D* and *G* can be interpreted as: one player is attempting to maximize its probability of winning, while the other player is trying to minimize the probability of winning of the first player. This is a sort of Minmax Game. The aforementioned interpretation begs the question of what exactly should be maximized or minimized. In fact, it is the Value Function (VF), as expressed below:

$$\min_G \max_D V(G,D) = E_{x \sim p_{data}}\left[\ln\left(D(x)\right)\right] + E_{z \sim p_z}\left[\ln\left(1 - D\left(G(z)\right)\right)\right]$$

From VF above, it can be observed that *G* seeks to minimize the expression, while *D* seeks to maximize it. Upon a closer look, one may easily realize that the expression above closely resembles the binary cross-entropy function, which is presented below (for one input):

$$L = -\sum y \ln \hat{y} + (1-y)\ln\left(1 - \hat{y}\right)$$

For the moment, let the negative sign and the summation be ignored. The remainder of the expression is just the binary cross-entropy function for a single input. *y* is the ground truth, i.e., the label, while $\hat{y}$ is the prediction of the label.

When $y = 1$, $\hat{y} = D(x) \Rightarrow L = \ln\left[D(x)\right]$

When $y = 0$, $\hat{y} = D\left(G(z)\right) \Rightarrow L = \ln\left[1 - D\left(G(z)\right)\right]$

Adding,

$$L = \ln\left[D(x)\right] + \ln\left[1 - D\left(G(Z)\right)\right]$$

It is worth understanding that this expression is valid for only 1 data point. It is necessary to extend it to the entire training dataset. To represent this mathematically, we need to use *expectations*. An expectation is the average value of the result of an experiment, if the experiment is performed a large value of times. The formula is straight forward:

$$E(x) = \sum x p(x)$$

It consists in adding the products of all possible outcomes and their respective probabilities. It is a sort of a weighted mean. So, applying the expectation on:

$$L = \ln\left[D(x)\right] + \ln\left[1 - D\left(G(Z)\right)\right]$$

We get:

$$E(L) = E\left(\ln\left[D(x)\right]\right) + E\left(\ln\left[1 - D\left(G(Z)\right)\right]\right)$$

We are adding all the scores with their probabilities:

$$\sum p_{data}(x)\ln\left[D(x)\right] + \sum p_z(z)\ln\left[1 - D\left(G(z)\right)\right]$$

But this is only true for discrete distributions. If we consider $\rho_{data}, \rho_z, \rho_g$ to be continuous distributions, we obtain:

$$\int p_{data}(x)\ln\left[D(x)\right]dx + \int p_z(z)\ln\left[1 - D\left(G(z)\right)\right]dz$$

The above integrals (sums) of probabilities are written in short form as , and so the VF for GANs is defined:

$$\min_G \max_D V(G,D) = E_{x \sim p_{data}}\left[\ln\left(D(x)\right)\right] + E_{z \sim p_z}\left[\ln\left(1 - D\left(G(z)\right)\right)\right]$$

In practice, it is necessary to explain how the GAN VF is optimized. In this case, the stochastic gradient descent method is considered. First, the big training loop (BTL) is entered. Within BTL, the learning of G is fixed. The inner loop for *D* (ILD) is then entered. ILD will continue for *k* steps. In IDL, *m* data points are sampled from the original data, and then *m* data points are sampled from the fake data. $\theta_d$ defined above is then updated by gradient descent:

$$\frac{\partial}{\partial \theta_d}\frac{1}{m}\left[\ln\left[D(x)\right] + \ln\left[1 - D\left(G(z)\right)\right]\right]$$

This is because the discriminator is trying to maximize the value function. After maximizing *k* updates of *D*, we exit ILD and turn to fixing the learning of *D*. Now, *G* has to be trained. For this, *m* data samples are sampled, and the weights of the generator $\theta_g$ are updated by gradient descent.

$$\frac{\partial}{\partial \theta_g}\frac{1}{m}\left[\ln\left[1 - D\left(G(z)\right)\right]\right]$$

This is because the generator is trying to minimize the value function. It is very important to recognize that for every *k* updates of the discriminator, the generator is updated once. A pertinent question remains – What is the guarantee that G would surely replicate $\rho_{data}(x)$ ? – it is necessary to prove that $\rho_g(x)$ will converge to $\rho_{data}(x)$ if G is able to find the global minimum of VF. In other words, it is necessary to show that the *global optimality p_g = p_data* at the global minimum of VF. So, for a fixed G in:

$$\mathrm{V}(G,D)=\int_x \ln\left[D(x)\right]+\rho_g(x)\ln\left[1-D(x)\right]dx$$

the aim is to find the value of *D* for which *D(x)* is maximum. It turns out that the answer is:

$$\frac{\rho_{data}(x)}{\rho_{data}(x)+\rho_g(x)}$$

Fixing *D(x)* as *D* above, then replacing *D(x)* with the expression, the expression for VF (with a fixed *D(x)*) becomes:

$$\min_{G}\mathrm{V}=E_{x\sim p_{data}}\ln\left(\frac{p_{data}(x)}{p_{data}(x)+p_g(x)}\right)$$
$$+E_{x\sim p_g}\ln\left(1-\frac{p_{data}(x)}{p_{data}(x)+p_g(x)}\right)$$

*or*

$$\min_{G}\mathrm{V}=E_{x\sim p_{data}}\ln\left(\frac{p_{data}(x)}{p_{data}(x)+p_g(x)}\right)$$
$$+E_{x\sim p_g}\ln\left(\frac{p_g(x)}{p_{data}(x)+p_g(x)}\right)$$

Recall that the burden of proof is defined as – *the probability distribution in the generated output is equal to the probability distribution in the original data*. So, it is logical to examine some of the methods used to measure the difference between two generations. One of the most famous methods is *JS divergence*. The formula for JS divergence looks surprisingly close to the above expression. JS divergence is expressed as:

$$JS\left(\rho_1\,||\,\rho_2\right)=\frac{1}{2}E_{x\sim p_1}\ln\left(\frac{\rho_1}{\frac{\rho_1+\rho_2}{2}}\right)+\frac{1}{2}E_{x\sim p_2}\ln\left(\frac{\rho_2}{\frac{\rho_1+\rho_2}{2}}\right)$$

Subtracting two logarithms from the JS divergence equation and using the result to modify the latest VF expression, the result becomes:

$$\min_{G}\mathrm{V}=E_{x\sim p_{data}}\ln\left(\frac{\rho_{data}(x)}{\frac{\rho_{data}(x)+\rho_g(x)}{2}}\right)$$
$$+E_{x\sim p_g}\ln\left(\frac{\rho_g(x)}{\frac{\rho_{data}(x)+\rho_g(x)}{2}}\right)-2\ln 2$$

Therefore,

$$\min_{G}\mathrm{V}=2JS\left(\rho_{data}\,||\,\rho_g\right)-2\ln 2$$

So, the minimum of the above expression (– 2ln2) is attained only when $\rho_{data}$ equals $\rho_g$. It has therefore been proven that at the global minimum of VF, $\rho_{data}$ equals $\rho_g$.

To summarize,

i. At the beginning: Here, $\rho_g$ does not 'know' what it is doing, so it does a bad job at mimicking $\rho_{data}$. The classifier discriminator is not classifying as well.
ii. After updating $\theta_{d:}$ Here, the classifier discriminator actually learns something, and can now distinguish between real and fake data.
iii. After updating $\theta_{g:}$ Here, the generator has learned something. $\rho_g$ is now closer to $\rho_{data}$. The classifier discriminator is trying to predict the true labels of the data point, but it is not performing as well.
iv. Finally: When the generator attains the minimum of the VF, then it has successfully replicated the distribution function of the original data points. $\rho_g$ is indistinguishable from $\rho_{data}$. It is now impossible for the discriminator to tell which data point is from the generated distribution and which data point is from the original distribution. So, the discriminator will output a half per input ( $D(x)=\frac{1}{2}$ ). This is a straightforward explanation of GANs. In practice, it is a difficult process to train.

Different types of GANs have been developed to enhance their capabilities, eight of which shall be discussed in this review – Conditional GAN, Cycle GAN, Deep Convolutional GAN (DC-GAN), Double GAN, Efficient GAN, Eigen GAN, Super Resolution GAN (SR-GAN) and Wasserstein GAN with Gradient Penalty (WGAN-GP).

Conditional Generative Adversarial Networks (cGANs) are an extension of the traditional GANs, designed to generate data conditioned on specific input labels or features. The architecture of a classical cGAN consists of two primary components: the generator and the discriminator each tailored to incorporate conditional information. The generator in a cGAN is responsible for producing synthetic data that resembles real data, but with the added capability of conditioning its output based on input labels. This is achieved by feeding both random noise and a conditional label into the generator. The random noise, typically sampled from a Gaussian distribution, serves as the latent variable, while the conditional label informs the generator about the specific type of data to produce. For instance, if the task is to generate images of handwritten digits, the generator would receive a label indicating which digit (0-9) to create. The architecture often employs deconvolutional layers to transform this combined input into high-dimensional image outputs. On the other hand, the discriminator functions as a binary classifier that evaluates whether a given input is real (from the training dataset) or fake (produced by the generator). In a cGAN, the discriminator also receives the conditional label as part of its input. This allows it to not only assess the authenticity of the image but also check if it corresponds correctly to the provided label. The discriminator typically consists of convolutional layers that progressively downsample the input image while extracting relevant features to make its classification. The training process of cGANs involves a min-max optimization framework where the generator aims to minimize its loss (i.e., successfully fooling the discriminator), while the discriminator seeks to maximize its accuracy in distinguishing real from fake images. This adversarial training continues iteratively, with both networks improving through feedback from one another. One significant advantage of cGANs over traditional GANs is their ability to control output generation based on specific conditions, leading to faster convergence during training and more relevant outputs during inference. For example, in applications such as image-to-image translation or text-to-image synthesis, cGANs can generate desired outputs by simply specifying conditions, making them versatile tools in various generative tasks (Mirza & Osindero, 2014; Isola *et al.*, 2018; DeVries *et al.*, 2019; Kinakh *et al.*, 2021; Boulahbal *et al.*, 2022; He *et al.*, 2022; Hou *et al.*, 2022; Kang *et al.*, 2024).

The Cycle Generative Adversarial Network (CycleGAN) architecture is a sophisticated model designed for image-to-image translation tasks, enabling the transformation of images from one domain to another without requiring paired examples. Introduced by (Zhu *et al.*, 2017), CycleGAN consists of two generator networks and two discriminator networks, each corresponding to two distinct image domains, often referred to as Domain-A and Domain-B. The architecture includes two generators: Generator-A and Generator-B. Generator-A is responsible for converting images from Domain-B to Domain-A, while Generator-B translates images from Domain-A to Domain-B. This bidirectional mapping is essential for the model's functionality, allowing it to learn the complex relationships between the two domains. Each generator operates under the principle of adversarial training, where its goal is to produce images that are indistinguishable from real images in the target domain, thereby fooling its corresponding discriminator. The discriminator networks, Discriminator-A and Discriminator-B, serve as binary classifiers that evaluate whether an image is real (from the actual dataset) or fake (produced by the generators). Discriminator-A assesses images generated by Generator-A against real images from Domain-A, while Discriminator-B does the same for images from Domain-B. This adversarial setup creates a competitive environment where the generators strive to improve their outputs while the discriminators enhance their ability to detect fakes. A defining feature of CycleGAN is its use of cycle consistency loss, which enforces that an image translated from one domain to another and then back again should closely resemble the original image. This is achieved through two processes: forward cycle consistency (Domain-B to Domain-A back to Domain-B) and backward cycle consistency (Domain-A to Domain-B back to Domain-A). The cycle consistency loss is calculated using L1 loss, ensuring that any distortions introduced during translation are minimized. In addition to cycle consistency loss, CycleGAN incorporates adversarial loss for both generators and discriminators. This loss encourages the generators to produce increasingly realistic images while pushing the discriminators to become more adept at distinguishing real from fake images. Furthermore, an identity loss can be applied, which helps maintain color composition between input and output images when an image from one domain is fed into its corresponding generator (Zhu *et al.*, 2017; Rao *et al.*, 2020; Sim *et al.*, 2020; Song & Ye, 2021; Rakhmatulin, 2022; Tadem, 2022; Torbunov *et al.*, 2022, 2023; An & Joo, 2023; Chen *et al.*, 2023; Choi *et al.*, 2023; Iacono & Khan, 2023; Sun *et al.*, 2023; Myers *et al.*, 2024; Wang & Lin, 2024).

DC-GAN is a sophisticated framework that enhances the traditional GAN by incorporating deep convolutional layers into both its generator and discriminator components. This design allows DCGANs to effectively generate high-quality images and has made them a popular choice in various applications, including image synthesis and data augmentation. At the core of the DCGAN architecture are two primary components: the Generator and the Discriminator. The Generator's role is to create images from random noise, while the Discriminator's task is to distinguish between real images from the training dataset and fake images produced by the Generator. This adversarial setup leads to a minimax game where the Generator aims to minimize the probability that its outputs are classified as fake, while the Discriminator seeks to maximize its accuracy in identifying real versus fake images. The Generator typically begins with a latent vector sampled from a normal distribution, which serves as input. This vector undergoes several transformations through layers of transposed convolutions, also known as deconvolutions, which upsample the input into larger feature maps. Each transposed convolution layer is usually followed by batch normalization to stabilize learning and improve convergence rates, along with ReLU activation functions that introduce non-linearity into the model. The final layer of the Generator employs a Tanh activation function, ensuring that the generated images are scaled to a range suitable for image data, typically between -1 and 1. The architecture is designed such that each layer progressively increases the spatial dimensions of the output until it reaches the desired image size. Conversely, the Discriminator is structured to classify images as real or fake. It takes an image input—either from the dataset or generated by the Generator—and processes it through multiple layers of convolutional operations. Unlike standard pooling layers, DCGANs utilize strided convolutions to reduce spatial dimensions while maintaining important features, enhancing the model's ability to learn discriminative patterns. The Discriminator also incorporates leaky ReLU activations, which allow for a small gradient when inputs are negative, thus preventing dead neurons during training. The final output of the Discriminator is passed through a Sigmoid activation function, producing a probability score indicating whether an input image is real (close to 1) or fake (close to 0). The training process for DCGANs involves alternating updates between the Generator and Discriminator. Initially, both networks are trained independently; however, they are updated in a way that each network's weights are adjusted based on its performance against the other. The Discriminator is trained on both real images (labeled as 1) and fake images (labeled as 0), using a binary cross-entropy loss function to gauge its performance. In contrast, during Generator training, it aims to produce outputs that maximize the Discriminator's probability of classifying them as real, effectively minimizing its own loss function (Radford *et al.*, 2016; Aslan *et al.*, 2019; Bolluyt & Comaniciu, 2019; Amyar *et al.*, 2020; Curtó *et al.*, 2020; Huang *et al.*, 2020; Venu, 2020; Mishra & Pathak, 2021; Rehm *et al.*, 2021; Blarr *et al.*, 2024).

A classical Double GAN extends the foundational GAN framework by integrating two adversarial pairs: a generator-discriminator pair for generating synthetic data and a secondary generator-discriminator pair that refines the output. This architecture is designed to enhance the quality and diversity of generated samples through a more complex interplay between the generators and discriminators. In a typical GAN, there are two main components: the Generator and the Discriminator. The Generator creates synthetic data from random noise, aiming to produce samples indistinguishable from real data. Conversely, the Discriminator evaluates both real and synthetic samples, classifying them as real or fake. The training process involves a min-max optimization game where the Generator seeks to minimize its loss (i.e., improve its ability to fool the Discriminator), while the Discriminator aims to maximize its accuracy in distinguishing real from fake samples. In a Double GAN architecture, this basic structure is augmented by introducing a second layer of generators and discriminators. The first generator produces initial synthetic data, while its corresponding discriminator assesses this output. The second generator takes the feedback from the first discriminator and generates refined samples, which are then evaluated by a second discriminator. This dual-layer approach allows for iterative refinement of generated outputs, facilitating higher fidelity and variability in the final results. The training dynamics in Double GANs are particularly intricate. Each generator strives to improve based on the feedback from its respective discriminator, creating a feedback loop that enhances learning efficiency. The architecture employs techniques such as batch normalization and specific activation functions (like Leaky ReLU for discriminators) to stabilize training and mitigate issues such as mode collapse, where the generator produces limited varieties of output. Moreover, Double GANs can incorporate additional mechanisms like conditional inputs or auxiliary losses to guide the generation process further. By conditioning on specific attributes or employing multi-task learning strategies, these networks can produce outputs that not only resemble real data but also conform to desired characteristics or styles (Shi *et al.*, 2021).

The architecture of a classical Eigen GAN is an innovative approach that integrates principles from linear algebra, specifically Singular Value Decomposition (SVD), into the framework of Generative Adversarial Networks (GANs). At its core, an Eigen GAN consists of two primary components: a generator and a discriminator, similar to traditional GAN architectures. However, what sets the Eigen GAN apart is its unique use of SVD in the generator's design. In a classical GAN, the generator typically learns to map random noise into a data distribution through a series of layers that transform this noise into realistic data samples. The Eigen GAN enhances this process by employing SVD to decompose the latent space effectively. This decomposition allows the generator to focus on the most significant features of the data distribution, thereby improving the quality and diversity of generated samples. Specifically, the generator utilizes the eigenvalues and eigenvectors derived from SVD to guide its learning process, which helps in capturing the underlying structure of the data more efficiently. The discriminator in an Eigen GAN operates similarly to that in traditional GANs, tasked with distinguishing between real and generated samples. However, it benefits from the enhanced representations produced by the generator's SVD-based architecture. The interaction between these two components follows the adversarial training paradigm: as the generator improves its ability to produce realistic samples, the discriminator simultaneously adapts to become better at identifying these samples as fake. Moreover, Eigen GANs often incorporate additional techniques such as layer-wise learning and regularization strategies to stabilize training and mitigate common issues like mode collapse. By focusing on eigenvalues during training, these models can achieve better convergence properties compared to standard GANs. This is particularly important given that traditional GANs are notorious for their training instability (He *et al.*, 2021; Kas *et al.*, 2024).

The architecture of a classical Efficient GAN is crafted to enhance both performance and computational efficiency. It comprises two primary components: the generator and the discriminator. The generator is responsible for producing realistic data samples from random noise, while the discriminator's job is to differentiate between real data samples and those created by the generator. To boost efficiency, Efficient GANs often utilize neural architecture search (NAS) to automatically identify the best network structures for both the generator and discriminator. This involves assessing multiple candidate architectures based on factors like model size, computational cost, and performance metrics such as Inception Score (IS) and Fréchet Inception Distance (FID). Furthermore, Efficient GANs may employ evolutionary algorithms to progressively refine the architectures, ensuring stability and high-quality outputs. By separating the optimization process into stages—first optimizing the generator with a fixed discriminator, then optimizing the discriminator with the best generator found—Efficient GANs strike a balance between computational efficiency and the capability to generate high-fidelity images (Gong *et al.*, 2024).

The architecture of a classical Super Resolution Generative Adversarial Network (SRGAN) is designed to enhance the resolution of low-quality images by generating high-resolution outputs that are perceptually realistic. At its core, the SRGAN consists of two main components: a Generator and a Discriminator, which work in an adversarial manner to improve image quality. The Generator is responsible for transforming low-resolution images into high-resolution counterparts. It typically employs a deep convolutional neural network (CNN) architecture that includes several layers of convolution, batch normalization, and activation functions like ReLU. The generator takes a low-resolution image as input and processes it through multiple convolutional layers, progressively increasing the spatial dimensions while refining features. This is often achieved through techniques such as residual learning, where skip connections are used to facilitate the flow of information and gradients, enhancing the model's ability to learn complex mappings from low to high resolution. In contrast, the Discriminator acts as a binary classifier that distinguishes between real high-resolution images and those generated by the Generator. It also employs a CNN architecture but is structured to output a single scalar value indicating whether the input image is real or fake. The Discriminator's role is crucial as it provides feedback to the Generator during training, pushing it to produce more realistic images by minimizing the adversarial loss. The training process involves optimizing two loss functions: the adversarial loss, which encourages the Generator to produce images that can fool the Discriminator, and the content loss, which ensures that the generated images maintain perceptual similarity to the original high-resolution images. Content loss is often computed using features extracted from pre-trained networks like VGG, focusing on higher-level representations rather than pixel-wise accuracy (Ledig *et al.*, 2016; Wang *et al.*, 2018, 2023b; Frizza *et al.*, 2022; Güemes *et al.*, 2022; Tian *et al.*, 2022; Baghel *et al.*, 2023; Huang & Omachi, 2023; Kuznedelev *et al.*, 2024).

Wasserstein GANs with Gradient Penalty (WGAN-GP) represent an advanced approach to training generative adversarial networks (GANs), addressing common challenges such as instability and poor convergence. The architecture consists of two primary components: the generator and the critic (often referred to as the discriminator in traditional GANs). The generator is responsible for creating synthetic data from random noise, while the critic evaluates the quality of both real and generated data. Unlike traditional GANs that minimize divergence metrics, WGAN-GP employs the Wasserstein loss, which approximates the Earth Mover's Distance. The goal is to maximize the difference in expected values between the critic's evaluation of real and generated samples, leading to a more stable training process. A key innovation in WGAN-GP is the introduction of a gradient penalty to enforce Lipschitz continuity, replacing the earlier method of weight clipping used in standard WGANs. The gradient penalty is implemented by penalizing the norm of the critic's gradients with respect to its inputs, ensuring that they remain close to 1. The penalty term effectively encourages smoother transitions in the critic's output, thus stabilizing training and preventing issues like vanishing or exploding gradients. Training dynamics in WGAN-GP typically require more iteration for the critic than for the generator. For every iteration of the generator, multiple iterations (often five) of the critic are performed. This allows the critic to better approximate the Wasserstein distance before updating the generator. The hyperparameter λλ, which controls the strength of the gradient penalty, is usually set around 10 based on empirical results (Chen & Tong, 2017; Gulrajani *et al.*, 2017; Kim *et al.*, 2018; Jolicoeur-Martineau & Mitliagkas, 2020; Kwon *et al.*, 2021; Milne & Nachman, 2022).

## Data Collection for Evaluating Generative Models in Plant Disease Identification

### *Inclusion criteria*

The inclusion criteria were designed to be comprehensive and specific to ensure the selection of recent, relevant, and high-quality studies. Firstly, the publication date criterion mandated that only articles published between 2018 and 2024, were considered, ensuring the inclusion of recent advancements. Secondly, the language criterion specified that only articles published in English were eligible. The relevance criterion focused on studies that applied generative models, such as GANs and VAEs, for plant disease diagnosis. The methodology criterion required robust experimental designs, including cross-validation and external validation where available, to ensure the reliability of the findings. The data type criterion specified that studies must use image data for plant disease diagnosis. Lastly, the outcome measures criterion included articles that report performance metrics like accuracy, precision, recall, F1-score, and ROC-AUC, among others.

### *Exclusion criteria*

The exclusion criteria were established to filter out studies that lacked relevance. Firstly, non-peer-reviewed studies were excluded to ensure that only rigorously vetted research was considered. This included excluding conference abstracts, editorials, and non-peer-reviewed articles. Secondly, studies with an irrelevant focus were excluded, specifically those not related to plant disease diagnosis or not using generative models. Thirdly, studies published in languages other than English were excluded to maintain consistency in language comprehension. Fourthly, duplicate studies were excluded to avoid redundancy and ensure that each study contributed unique data. Lastly, studies that did not provide sufficient data or details on methodology and results were excluded under the criterion of insufficient data.

### *Data extraction and management*

Data extraction and management followed a structured approach for collecting and handling data from the selected studies. It began with bibliographic information, including details such as the author(s), title, journal, and year of publication, ensuring accurate recording of all relevant citation information. Next, it specified study characteristics, detailing the study design, sample size, type of plant diseases studied, and the generative models used, which helped in understanding the context and scope of each study. The section also covered data characteristics, involving the type of data used (e.g., leaf images, whole plant images), data sources, and preprocessing techniques, ensuring the data was appropriately categorized and prepared for analysis. Additionally, it included model details, such as the architecture of the generative models, training parameters, and validation methods, providing insight into the technical aspects of the studies. Information on performance metrics, including accuracy, precision, recall, F1-score, AUC-ROC, and computational efficiency, which are crucial for evaluating the effectiveness of the models, was collected. The final section comprised key findings, summarizing the main results, conclusions, limitations, and future research directions, ensuring a comprehensive understanding of each study's contributions and gaps.

*Analysis*

In addition to reporting the observations from each section, comparative analyses were conducted to assess how various factors, such as dataset size and model training parameters, inter alia, contributed to model performance.

## RESULTS AND DISCUSSION

### Co-authorship Analysis

The co-authorship network visualization in Figure 1a (and its extension in Figure 1b) provides a look into the collaborative relationships among researchers. Chen, J appears to be a central figure in this network, with multiple connections to other authors such as Zhang, D, Wang, Z, and Liu, H. This indicates that Chen, J has collaborated with several other researchers, suggesting a significant role in the research community. The network shows distinct clusters of authors who frequently collaborate. For example, Li, D, Tang, Y, and Liu, H form a cluster, indicating a close-knit group of researchers working together. Another cluster includes Zhao, Y, Xu, X, and Li, Y, which also suggests a strong collaborative relationship among these authors. The color gradient provides insights into the temporal trends of co-authorship. For instance, collaborations involving Wang, Y and Wang, Q are more recent (closer to 2023), as indicated by the yellow color. Earlier collaborations (closer to 2021) are represented by darker blue nodes and edges, such as those involving Li, D and Liu, H. The network is relatively interconnected, with several authors having multiple co-authorship links. This interconnectedness suggests a collaborative research environment where knowledge and ideas are shared among various researchers. Authors with multiple connections, such as Chen, J and Zhao, Y, could be considered research hubs or key influencers in their respective fields. Their extensive collaboration network might indicate their leadership in research projects or their role in facilitating collaborations among other researchers.

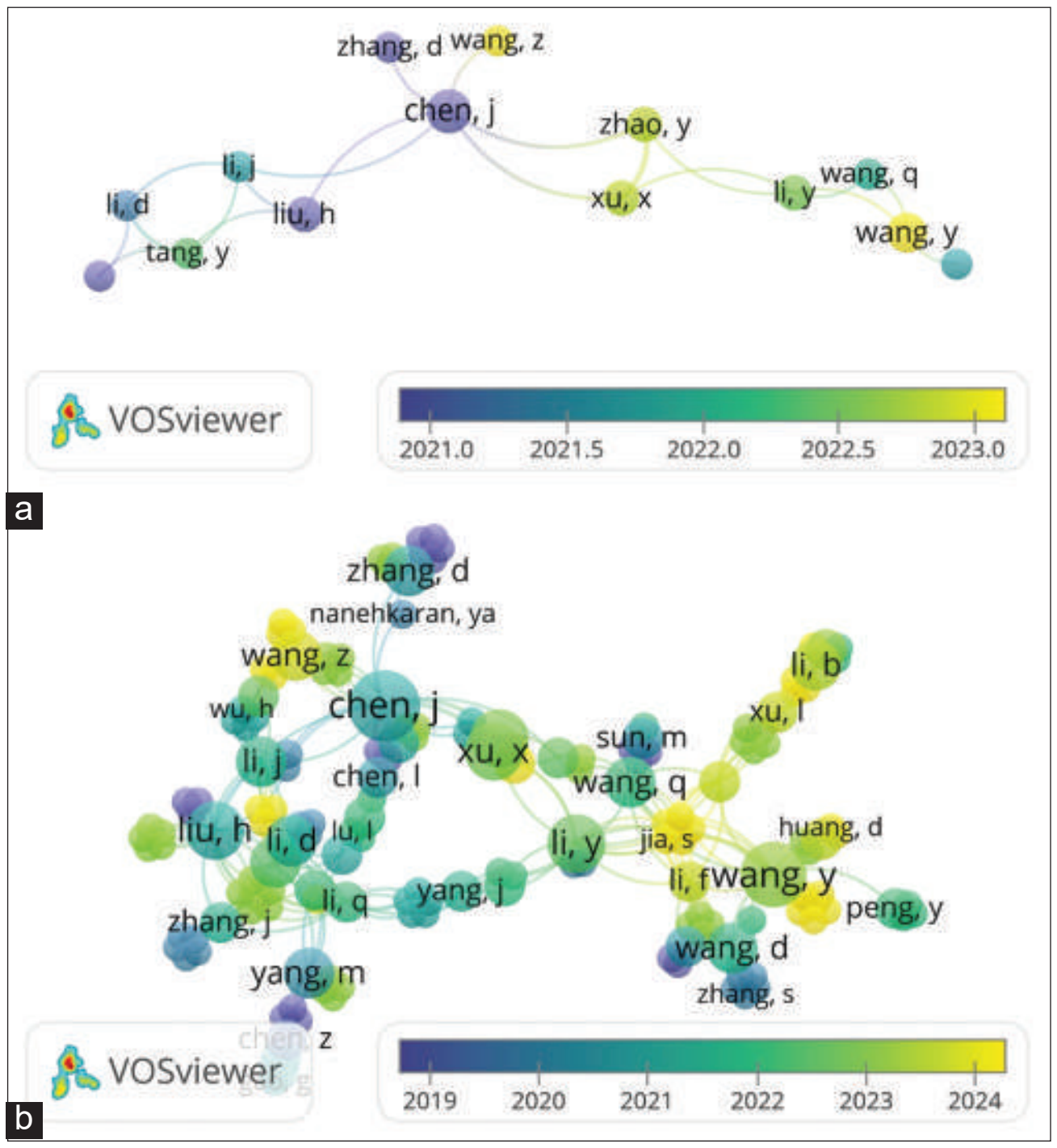

**Figure 1:** Co-authorship networks. a) Co-authorship network for authors having 3 or more publications and b) Co-authorship network for authors having 1 or more publications

### Keyword Co-occurrence Analysis

The keyword co-occurrence network visualization (Figure 2) provides an overview of the relationships between various keywords in the context of this review. In this network, keywords are represented as nodes, with the size of each node indicating the frequency of the keyword's occurrence. The lines connecting the nodes represent the co-occurrence of keywords in the same documents, with the thickness of the lines indicating the strength of the co-occurrence. The most prominent keyword in the network is "deep learning," which is central and has the largest node, indicating its high frequency and importance in the research domain. Other significant keywords include "convolutional neural networks," "learning systems," and "disease detection," which are closely related to deep learning. The network also includes more specific terms such as "tomato leaf," "leaf disease detection," and "plant leaf," highlighting the focus on plant disease identification. The color of the nodes and edges represents the average publication year of the documents in which the keywords appear, with a gradient from blue (earlier years) to yellow (more recent years). This color coding provides insight into the temporal evolution of the research topics. For example, keywords like "deep learning" and "convolutional neural networks" are more recent, as indicated by their yellowish color, while some other terms might appear in earlier years, shown in blue or green. This keyword co-occurrence network is interesting and relevant because it provides a visual representation of the research landscape, showing how different topics are interconnected and how the focus of research has evolved over time. It helps researchers identify key areas of interest, emerging trends, and potential gaps in the literature.

### Author-based Citation Analysis

This report assesses researchers' impact through citation metrics, highlighting their contributions to the academic community. Notably, the collaborative work of Njuki, S., Too, E.

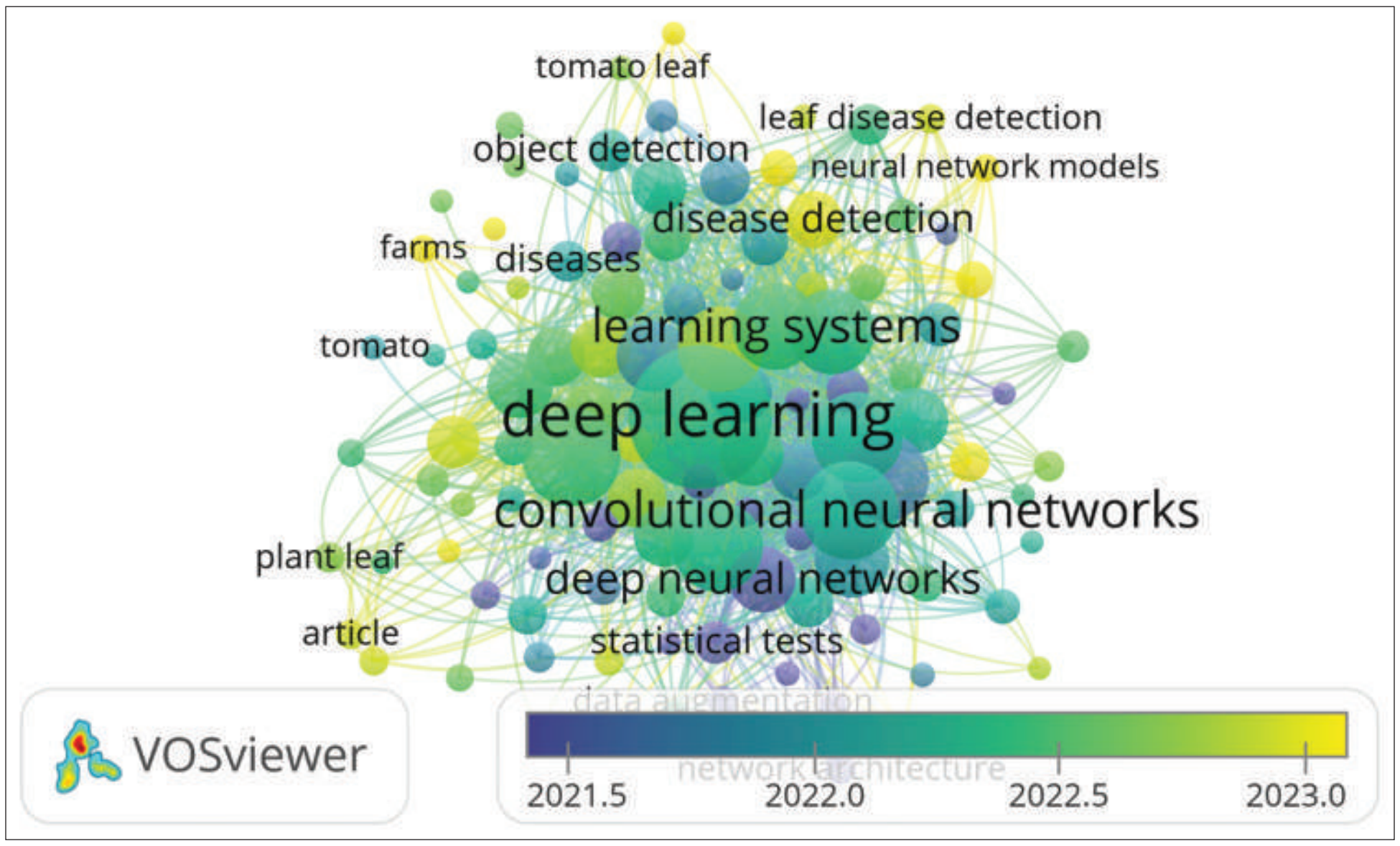

**Figure 2:** Keyword co-occurrence network.

C., Yingchun, L., and Yujian, L. stands out with a remarkable total of 764 citations. This substantial figure indicates a significant impact on their field, suggesting that their research addresses critical issues or introduces innovative concepts that resonate widely within the academic community. Following closely, Arnal Barbedo's document with 485 citations also reflects a significant impact. Additionally, the collaborative work of authors such as Ali, M. E., Apon, S. H., Arko, P. S., Iqbal Khan, M. A., Nowrin, F., Rahman, C. R., and Wasif, A. has garnered 356 citations, further emphasizing their contributions to critical issues in their respective fields. Zhang's contributions are noteworthy due to the accumulation of 283 citations across three separate documents. Similarly, Chen's six documents totaling 261 citations highlight a prolific output and sustained influence in his research area. The work co-authored by Coppola, G., Hu, Y., Liang, Q., Sun, W., and Xiang, S. has received 200 citations, indicating valuable contributions that advance knowledge within their disciplines. Authors Biswas, D. and Mukti, I. Z.'s co-authored document with 190 citations further illustrates the significance of their research contributions. The analysis reveals that citation counts serve as a valuable metric for assessing the impact of researchers' work within their fields. The significant citation numbers achieved by Njuki, Arnal Barbedo, and others illustrate how addressing critical issues or introducing innovative concepts can lead to widespread recognition and influence in academia. Moreover, the presence of multiple documents by authors like Zhang and Chen indicates not only prolific output but also an established reputation within their respective areas of study. Such patterns suggest that these researchers are not only contributing valuable knowledge but are also shaping ongoing discourse in their fields. The findings highlight the importance of collaboration among researchers to enhance impact through collective expertise and diverse perspectives. Future research can benefit from examining the specific themes or methodologies employed by these highly cited works to understand better what factors contribute to high citation counts. Additionally, it may be beneficial to explore how these citation metrics correlate with practical applications or advancements in technology and policy influenced by this research.

## Country-based Citation Analysis

The country-based citation network (Figure 3) provides a fascinating insight into the global landscape of academic and research collaborations. This network map illustrates the relationships and citation links between different countries. Each node represents a country, and the size of the node indicates the volume of citations or collaborations associated with that country. The lines connecting the nodes represent the citation links between the countries, with the thickness of the lines indicating the strength or frequency of these links. China and India are the two most prominent nodes in this network, suggesting that they have a significant number of citations or collaborations with other countries. China's central position and its connections to several countries, including Turkey, Canada, Brazil, Algeria, and Brunei Darussalam, indicate that it plays a crucial role in the global citation network. This central position suggests that China acts as a major hub for academic or research collaborations, facilitating the exchange of knowledge and ideas across borders. India, another significant node, is connected to Saudi Arabia and also shares links with China. The connection between India and China is particularly noteworthy, as it suggests a strong collaborative relationship between these two countries in terms of citations or research output. This relationship highlights the importance of regional collaborations in advancing scientific research and innovation. The visualization of this network is not only interesting but also highly relevant. It provides valuable insights into the interconnectedness of the global research community and underscores the importance of international collaborations in advancing knowledge and innovation in deep learning for plant disease identification.

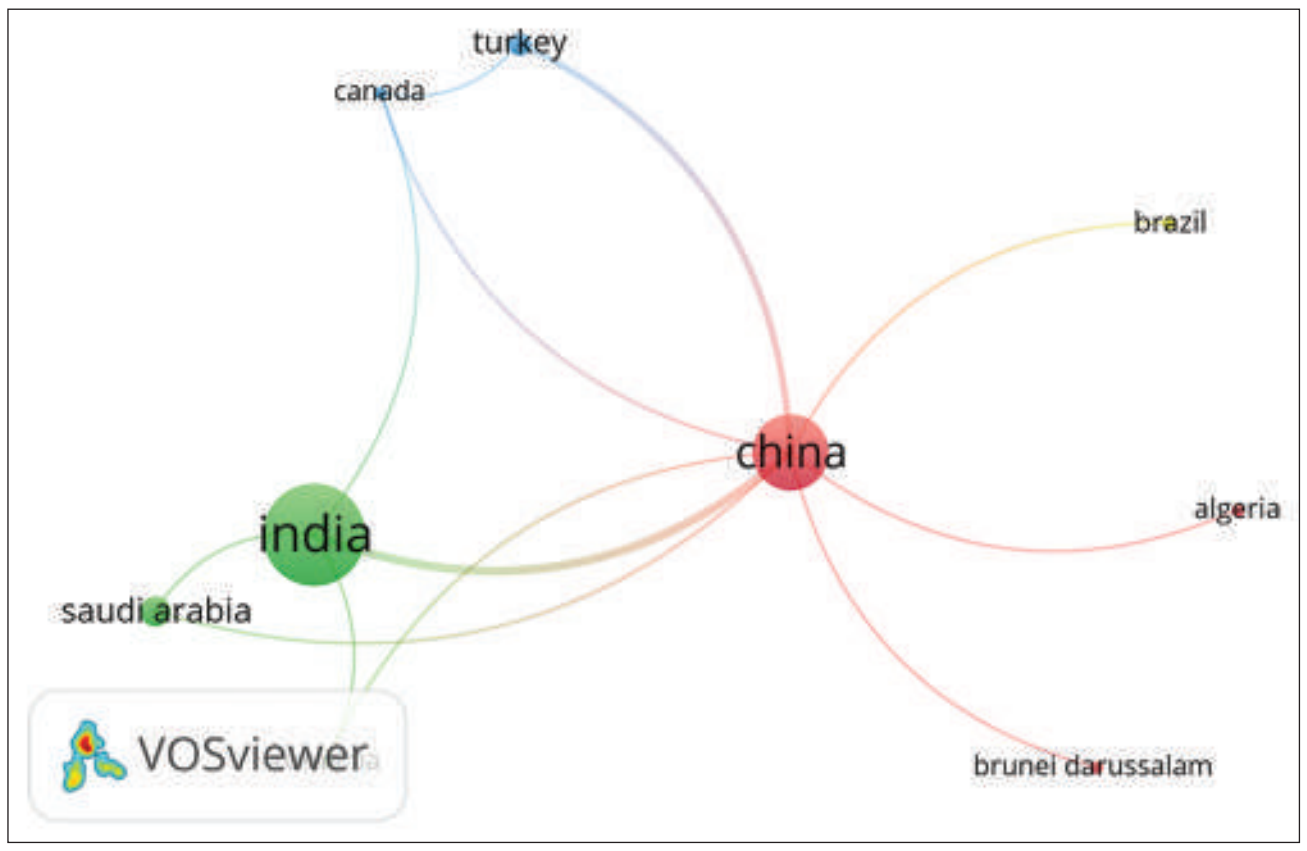

**Figure 3:** Country-based citation network

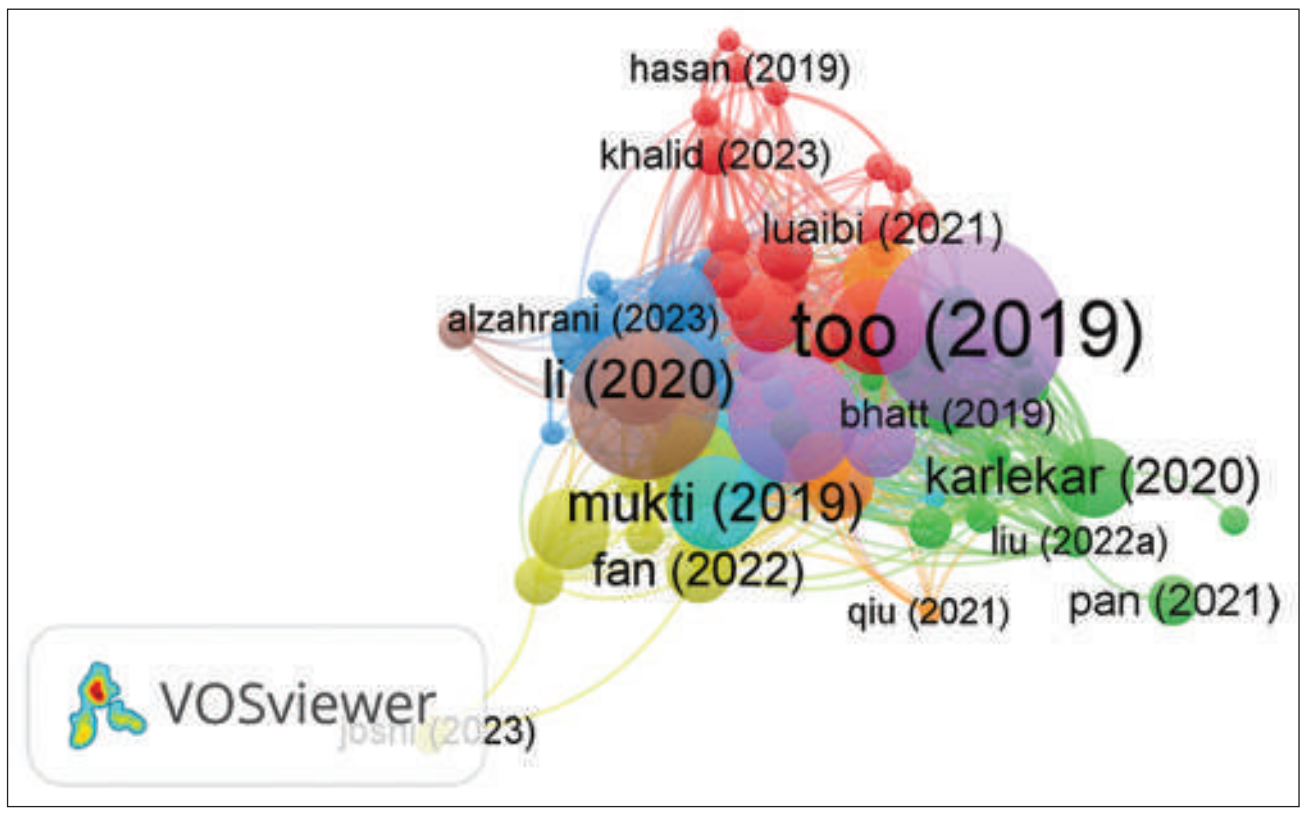

**Figure 4:** Document-based bibliographic coupling network

## Document-based Bibliographic Coupling

The document-based bibliographic coupling network (Figure 4) provides a comprehensive overview of the relationships between various scientific publications based on their shared references. This network highlights the interconnectedness of research papers and the influence of key documents within a specific field. Each node in the network represents a document, with the size of the node indicating the number of citations the document has received. The edges connecting the nodes represent the strength of the bibliographic coupling, which is determined by the number of shared references between the documents. One of the most prominent nodes in this network is "too (2019)," which suggests that this document has a significant number of citations and strong bibliographic coupling with other documents. This prominence indicates that "too (2019)" is a highly influential paper within its research domain, serving as a foundational reference for subsequent studies. Other notable nodes include "mukti (2019)," "karlekar (2020)," "li (2020)," and "fan (2022)," each of which also plays a crucial role in the network. The presence of these nodes highlights the key contributions of these documents to the field and their impact on the development of related research. The network is divided into clusters, each represented by different colors, indicating groups of documents that are closely related in terms of their bibliographic references. These clusters reveal the thematic structure of the research field, showing how different topics or subfields are interconnected. For example, documents within the same cluster are likely to address similar research questions or methodologies, while connections between clusters indicate interdisciplinary collaborations or the integration of diverse perspectives. This visualization is particularly valuable for researchers, as it provides insights into the structure of the research field, highlighting influential documents and the relationships between them. By examining the network, researchers can identify key works that have shaped the field, understand the development of specific research topics, and find potential collaborators who are working on related issues. Additionally, the network can help researchers identify gaps in the literature, guiding future research efforts to address these areas.

## Author-based Co-citation Analysis

The author-based co-citation network (Figure 5) is a visualization of the relationships between authors based on the frequency with which they are cited together in academic literature. Each node in the network represents an author, and the size of the node indicates the number of co-citations that author has received. The lines connecting the nodes represent the co-citation links between authors, with the thickness of the lines indicating the strength of the co-citation relationship. The network is color-coded to represent different clusters of authors who are frequently co-cited together. For example, the red cluster includes authors such as Zhang Y., Liu B., and Chen Y., indicating that these authors are often cited together in the same papers. Similarly, the green cluster includes authors like He K., Sun J., and Szegedy C., while the blue cluster includes authors like Salathe M., Hughes D. P., and Chen J. The yellow cluster includes authors such as Sharif M., Zhang S., and Khan M. A. This visualization is interesting and relevant because it provides insights into the structure of academic research communities and the relationships between influential authors. By examining the clusters and connections, researchers can identify key authors, influential research groups, and emerging trends in a particular field. Additionally, the network can help researchers understand the intellectual structure of a research domain and identify potential collaborators or influential works that they may have overlooked.

## Publication and Citation Trends

The publication and citation trends (Figure 6) in this context of deep learning applications in plant disease identification reveal insights into the evolution of this research field. The publication trend graph shows a steady increase in the number of publications from 2018 to 2024, with a notable peak in 2023. This upward trajectory indicates a growing interest and investment in leveraging deep learning for plant disease identification, reflecting advancements in technology and the increasing importance of sustainable agriculture. The significant jump between 2021 and 2022 suggests a surge in research activities, possibly driven by new funding opportunities, technological breakthroughs, or heightened global interest

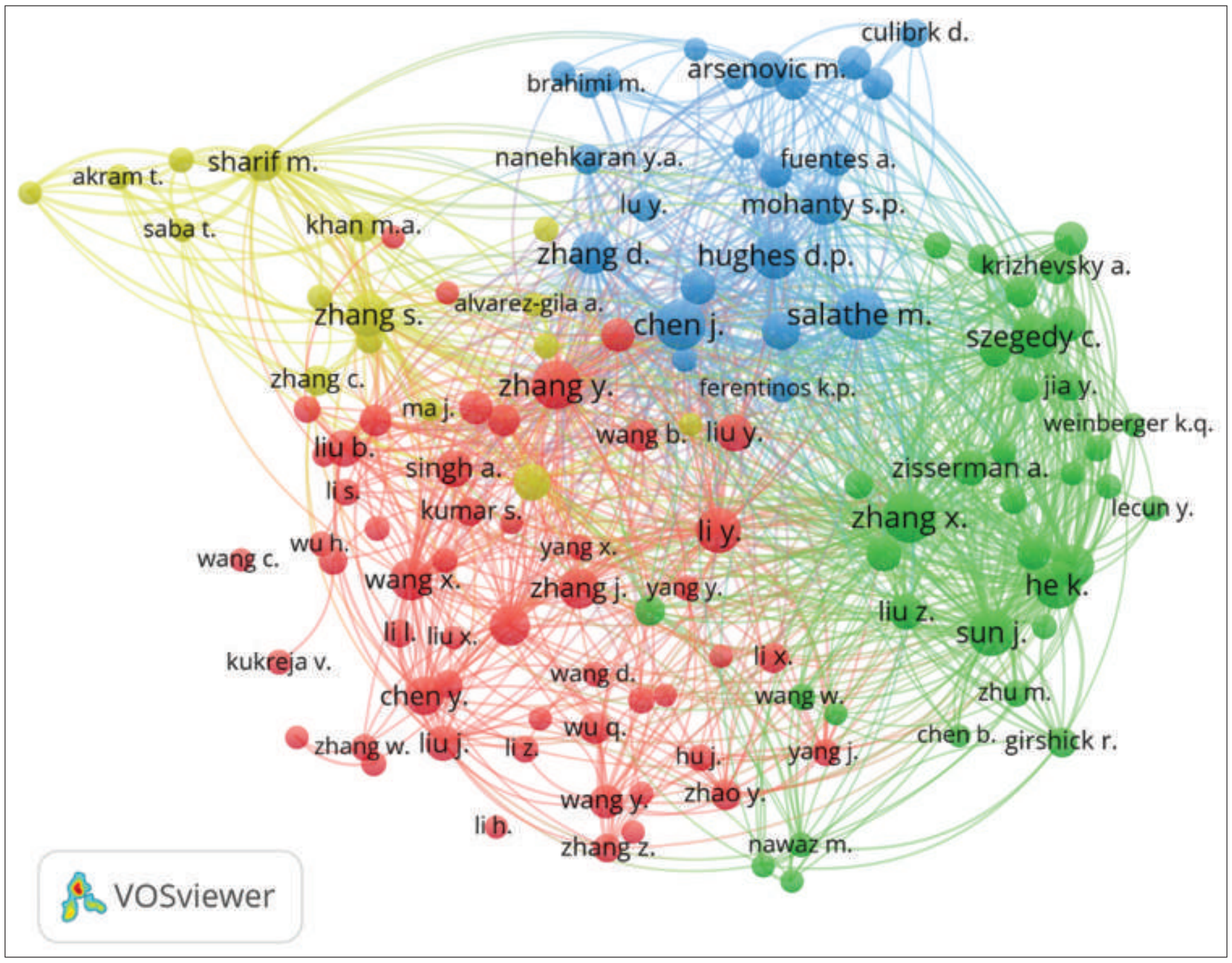

**Figure 5:** Author-based co-citation network

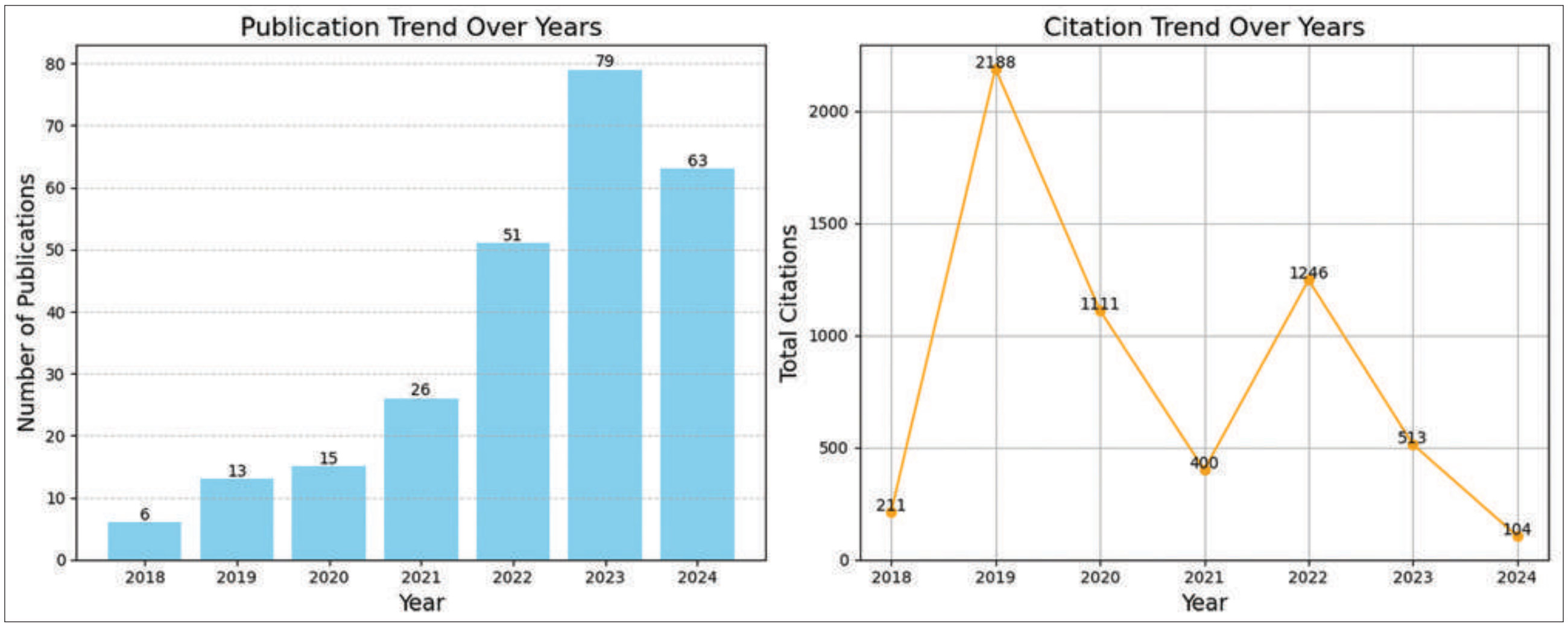

**Figure 6:** Publication and citation trends

in addressing plant health challenges. On the other hand, the citation trend graph presents a more fluctuating pattern. Citations peaked dramatically in 2019, followed by a decline in the subsequent years, with another rise in 2022 before dropping again in 2023 and 2024. This fluctuation suggests that while the number of publications has generally increased, the impact or recognition of these publications, as measured by citations, has varied significantly year by year. The peak in citations in 2019, despite a lower number of publications, indicates that earlier research in deep learning applications for plant disease identification might have had a higher impact or was more widely recognized. Conversely, the decline in citations in 2021, despite an increase in publications, could indicate that newer research is either not yet widely recognized or cited, or it may not be as impactful as earlier works. The relationship between the two trends is particularly interesting. Despite the increasing number of publications, the citation count does not follow a consistent upward trend. This discrepancy highlights the complex nature of academic impact, where the quantity of research output does not necessarily correlate with its quality or influence. The rise in citations in 2022 aligns with an increase in publications, suggesting a possible correlation between the two for that year. However, the subsequent decline in citations in 2023 and 2024, despite high publication numbers, indicates

**Table 1: Data-focused summary table of generative model applications for plant disease identification**

| Reference | Plant diseases studied | Sources of image data | Data size/Train-Test splitting | Preprocessing techniques |
|---|---|---|---|---|
| Alshammari *et al.*, 2024 | Leaf spot, leaf curl and slug damage in pear | Sardegna, Italy. Honor 6x smartphone and DSLR camera (Private) | 3,505 images (3,006 leaf images and 499 fruit images). Train –Validation – Test split=70% – 10% – 20% | - |
| Bi & Hu, 2020 | 38 disease classes of Fungal, Bacterial, Viral and Mite pest origins on 14 crop species – Apple, Blueberry, Cherry, Corn, Grape, Orange, Peach, Bell Pepper, Potato, Raspberry, Soybean, Squash, Strawberry and Tomato | Kaggle and PlantVillage | 4,384 images – 873 for training and the remainder for testing | All images were resized to 128 x 128 pixels. Pixel values were normalized. Data augmentation via geometric transformation including vertical and horizontal flipping, rotation and brightness adjustments |
| Förster *et al.*, 2019 | Leaf powdery mildew in barley plants | Hyperspectral Imaging (HIS) microscope, with spectral resolution up to 2.3nm across 420nm – 830nm wavelength range (Private) | 5,250 image patches – 3,500 from healthy leaves and 1,750 from inoculated leaves. 7,000 more patches were obtained using mirroring techniques. Training dataset=patches from 4-10 days after inoculation (DAI). Testing dataset=patches from 5-11 DAI. | Noisy spectral bands were removed from images. Image patches were extracted and resize to 100 x 100 pixels. The diversity of the training dataset was accomplished through augmentation techniques such as mirroring. |
| Krishnakumar & Balasubrahmanyan, 2024 | Leaf bacterial spot, early blight, late blight in tomato, inter alia. Bacterial spot in Bell Pepper. | PlantVillage | 54,303 images. | The images were resized – reduced to 512 x 512 pixels. Image denoising was achieved with Hybrid Fourier Filter Denoising (HFFD) and augmented with EigenGAN |
| Lopes *et al.*, 2023 | Multispectral whole-plant images of wheat yellow rust | UAV flights conducted over experimental plots (Private) | 700 plots, including 592 annotated plots – 430 plots classified as mild (72.6%), 106 as unhealthy (17.9%) and 56 as healthy (9.45%) | The remaining images were resized to 128 x 128 x 5. Mild samples were excluded from model training. 47 indices were extracted from the imagery using a vegetation index extractor. |
| Van Marrewijk *et al.*, 2022 | Leaf scab on apple trees | Plant Pathology 2020 dataset (fgvc7) Plant Pathology 2021 dataset (fgvc8) | *Training*: Plant Pathology 2020 dataset (fgvc7) *Testing*: Plant Pathology 2021 dataset (fgvc8) | - |
| Min *et al.*, 2023 | Leaf black rot, leaf scab and leaf rust for Apple; Leaf early blight and leaf late blight for potatoes; Leaf esca and leaf blight for grapes. | PlantVillage | Total=4,750 samples. 1,775 samples for apple; 1,923 samples for grape; 1,152 samples for potato. | - |
| Miranda *et al.*, 2022 | Physical damage and disease in grapevine berries | Kühn-Institut Geilweilerhof in Siebeldingen, Germany, using a field phenotyping platform called Phenoliner. (Private) | 616 images. Train-Test split=80%-20% | Non-overlapping 130 x 130-pixel patches were extracted from the images and further resized to 64 x 64 pixels to reduce computational cost. |
| Shete *et al.*, 2020 | Maize tassel phenotyping | Singapore Whole sky IMaging SEGmentation (SWIMSEG) Database for foreground images; HYTA dataset for background images. | 218 images 130 (65%) for training. 88 (35%) for testing. | Image cropping Image segmentation Image resizing |

(*Contd...*)

Table 1: (*Continued*)

| Reference | Plant diseases studied | Sources of image data | Data size/Train-Test splitting | Preprocessing techniques |
|---|---|---|---|---|
| Sun *et al.*, 2020 | Leaf canker disease on citrus | Nikon D7500 Camera (Private) | Total=3,152 images. Train-Validation-Test splitting=2000 images – 500 images – 652 images. | Cropping lesion areas from leaf images, Applying edge smoothing, Using image pyramids to generate image lesions of various sizes (augmentation). |
| Vasudevan & Karthick, 2022 | Leaf black measles, black rot and late blight on Grapes | Two sources: 1) real-time images captured using an SD1000 camera in Tamilnadu, India 2) PlantVillage | 2,500 images=500 real-time images and 2,000 PlantVillage images. Healthy-images=1000 Disease-images=1,500 | Data augmentation via: 1) Geometric transformations – flipping and translating. 2) Graph-based segmentation to accurately extract leaf areas from images. |
| Wang & Cao, 2023 | 24 types of disease in the PlantVillage dataset; 27 types of diseases in the AI Challenger 2018 dataset. | PlantVillage; AI Challenger 2018. | 54,306 images in PlantVillage. 36,379 images in AI Challenger 2018. Train-Test split=80% - 20% | Images were resized. |
| Wu *et al.*, 2020 | Leaf late blight water mold, septoria fungus, target spot bacterium and YLCV on tomato | PlantVillage | Total=1,500 images. Each disease class contained 300 images. The healthy class also contained 300 images. | All images were resized to the input dimensions required by the neural networks (224×224 pixels for GoogLeNet, AlexNet, and ResNet; 299×299 pixels for VGG), then converted to RGB color scale and saved in JPG format. |
| Zhang *et al.*, 2023 | Fusarium wilt on Cotton | College of Agriculture at Shihezi University | Training Set: 154 images (71 healthy, 83 diseased) Validation Set: 78 images (36 healthy, 42 diseased) Test Set: 185 images (118 healthy, 67 diseased) | Image resolution was compressed from 569×569 pixels to 256×256 pixels. |
| Zhao *et al.*, 2022 | Leaf bacterial spot, early blight, late blight, mold, partial leaf spot, mosaic and YLCV on Tomato. Other diseases from Apple, Maize, Grape and Potato. | PlantVillage | 31,361 images. 80% for training, 20% for testing. | Geometric transformations, including flipping and translating. |
| Zilvan *et al.*, 2019 | For corn, four disease classes were examined: Corn Gray Leaf Spot, Corn Common Rust, Corn Northern Leaf Blight, and a healthy class. For potato, three disease classes were analyzed: Potato Early Blight, Potato Late Blight, and a healthy class. | PlantVillage | Corn=3,852 images Potato=2,152 images Train-Validation-Test=80% - 10% - 10% | The images were resized from their original dimensions of 256 x 256 pixels to 64 x 64 pixels to standardize input size for the deep learning models. Additionally, salt and pepper noise was applied to the images to simulate real-world conditions and test the robustness of the proposed system against impulse noise. |

that the newer publications have not yet garnered significant citations.

## Generative Modeling for Plant Disease Identification

The studies encompass a wide range of plant diseases (Table 1), demonstrating significant diversity. Alshammari *et al.* (2024) focused on leaf spot, leaf curl, and slug damage in pear. Bi and Hu (2020) examined 38 disease classes of fungal, bacterial, viral, and mite pest origins on 14 crop species, including apple, blueberry, cherry, corn, grape, orange, peach, bell pepper, potato, raspberry, soybean, squash, strawberry, and tomato. Förster *et al.* (2019) studied leaf powdery mildew in barley plants. Krishnakumar and Balasubrahmanyan (2024) investigated leaf bacterial spot, early blight, late blight in tomato, and bacterial spot in bell pepper. Lopes *et al.* (2023) researched wheat yellow rust, while van

**Table 2: Model-focused summary table of generative model applications for plant disease identification**

| Reference | Model | Training parameters | Performance metrics results |
|---|---|---|---|
| Alshammari *et al.*, 2024 | CycleGAN | / | *Accuracy*: ResNet50=86.59%, VGG19=84.28% EfficientNetB0=73.48%<br>*Precision*: ResNet50=85.17%, VGG19=85.47% EfficientNetB0=71.42%<br>*Recall*: ResNet50=91.38%, VGG19=88.38% EfficientNetB0=74.70%<br>*F1-Score*: ResNet50=87.73%, VGG19=86.84% EfficientNetB0=71.84% |
| Bi & Hu, 2020 | WGAN-GP | Epochs=700<br>Optimizer=RMSprop<br>Learning rate=0.0001<br>Batch size=100<br>Loss function=Wasserstein distance<br>Overfitting mitigation technique=LSR | Recall, Precision and F1-Score<br>*Experiment I*:<br>F1-Score=0.46<br>*Experiment II*:<br>F1-Score=0.76<br>*Experiment III*:<br>F1-Score=0.78<br>*Experiment IV*:<br>F1-Score=0.81 |
| Förster *et al.*, 2019 | CycleGAN | Loss functions included:<br>1) adversarial loss,<br>2) cycle-consistent loss. | Visual evaluation of the generated images=praiseworthy |
| Krishnakumar & Balasubrahmanyan, 2024 | EigenGAN | Leaning rate,<br>batch size and<br>number of epochs. | Tomato bacterial spot RMSE (to quantify the difference between the original and denoised images) = 0.32702<br>Peak signal to noise ratio (PSNR) for bacterial spot in bell pepper HFFD denoising=58.7179. |
| Lopes *et al.*, 2023 | PlantPlotGAN<br>DC-GN<br>WGAN<br>VQGAN | Optimizer=Adam<br>Learning rate=2e-4<br>Adam Optimizer settings of β1=0.5, β2=0.99<br>Epochs=50<br>Training time=14 hours | *Evaluation metrics*:<br>Fréchet Inception Distance<br>Chi-square test<br>Information Coefficient<br>Bhattacharyya Coefficient<br>*Results*:<br>PlantPlotGAN outperformed the rest |
| Van Marrewijk *et al.*, 2022 | CycleGAN | Learning rate,<br>Batch size,<br>Number of epochs,<br>Generator loss function,<br>Discriminator loss function | *For the Benchmark dataset*:<br>Precision: 0.99, Recall: 0.38, F1 Score: 0.55, Accuracy: 68.43%<br>*For the ExtraTrain dataset*:<br>Precision: 0.99, Recall: 0.05, F1 Score: 0.55, Accuracy: 68.43%<br>*For the ExtraValid dataset*:<br>Precision: 0.99, Recall: 0.41, F1 Score: 0.58, Accuracy: 69.49%<br>*For the ExtraBenchmark dataset*:<br>Precision: 0.99, Recall: 0.53, F1 Score: 0.69, Accuracy: 77.49% |
| Min *et al.*, 2023 | CycleGAN | 1) Iterations=100,000<br>2) Adam Optimizer settings of β1=0.5, β2=0.999<br>3 & 4) Generator and Discriminator learning rates=0.0002<br>5) Loss function was a combination of:<br>- Adversarial loss,<br>- Cycle-consistency loss,<br>- Identity loss<br>- Background loss,<br>Reduction ratio | Potato Leaf Classification:<br>*ResNet 18*:<br>Imbalance: Precision 0.9768, Recall 0.9743, F1 Score 0.9755<br>CycleGAN: Precision 0.9791, Recall 0.9762, F1 Score 0.9774<br>Proposed: Precision 0.9793, Recall 0.9787, F1 Score 0.9790<br>*DenseNet 161*:<br>Imbalance: Precision 0.9879, Recall 0.9859, F1 Score 0.9867<br>CycleGAN: Precision 0.9912, Recall 0.9899, F1 Score 0.9905<br>Proposed: Precision 0.9903, Recall 0.9926, F1 Score 0.9914<br>*MobileNet v2*:<br>Imbalance: Precision 0.9855, Recall 0.9852, F1 Score 0.9853<br>CycleGAN: Precision 0.9868, Recall 0.9845, F1 Score 0.9856<br>Proposed: Precision 0.9882, Recall 0.9888, F1 Score 0.9885<br>*EfficientNet b0*:<br>Imbalance: Precision 0.9969, Recall 0.9968, F1 Score 0.9968<br>CycleGAN: Precision 0.9996, Recall 0.9985, F1 Score 0.9991<br>Proposed: Precision 0.9998, Recall 0.9992, F1 Score 0.9995 |

(*Contd...*)

Table 2: (*Continued*)

| Reference | Model | Training parameters | Performance metrics results |
|---|---|---|---|
| Miranda *et al.*, 2022 | VAE | Bach size=64<br>Optimizer=Adam<br>Initial learning rate=0.0005<br>Early stopping<br>Loss functions – Reconstruction loss and KL divergence | *Accuracy*=92.3% |
| Shete *et al.*, 2020 | DC-GAN<br>Sky background generation model | *DC-GAN*:<br>Epochs=2,500+Learning rate=1.5e-5<br>Optimizer=Adam<br>Loss=cross-entropy<br>*Sky background generation model*:<br>Epochs=600<br>Optimizer=Adam<br>Loss=cross-entropy | *Inter-annotator agreement error (IAAE) scores*:<br>For top 5 annotators, IAAE=0.222<br>*Classification accuracy*:<br>72.36% of real images were classified as real, while 60.62% of generated images were classified as generated. |
| Sun *et al.*, 2020 | Combination of WGAN-GP & Conditional GAN (CiGAN) | Learning rate<br>Batch size<br>Number of training epochs | *Using the bilinear interpolation method*:<br>Inception Score=6.12±0.08<br>Fréchet Inception Distance=20.02±1.04 |
| Vasudevan & Karthick, 2022 | Efficient GAN (E-GAN) | Batch size=32<br>Learning rate=0.0001<br>Optimizer=Adam<br>Epochs=50<br>Activation=ReLU<br>Loss=cross-entropy | Image evaluation was performed using CapsNet model.<br>CapsNet *Accuracy*=97.63% |
| Wang & Cao, 2023 | GACN.<br>The classification component of the GACN was based on ResNet18. | *General*:<br>Batch size=128,<br>Epochs=300<br>*Generator*:<br>optimizer=Adam, with β1=0.5, β2=0.999<br>learning rate=0.0001<br>*Discriminator*:<br>optimizer=Adam, with β1=0.5, β2=0.999<br>learning rate=0.0004<br>*ResNet18*:<br>learning rate=0.0004<br>weight decay=1e-4 | *GACN ACCURACY measures*:<br>PlantVillage dataset=99.78%<br>AI Challenger 2018=86.52%<br>*GACN ACCURACY measures with 5-fold cross-validation*:<br>PlantVillage dataset=99.63%<br>AI Challenger 2018=86.37%<br>*Other metrics*:<br>Peak Signal-to-Noise Ratio (PSNR),<br>Structural Similarity Index (SSIM),<br>Perceptual Image Quality Evaluator (PIQE),<br>Natural Image Quality Evaluator (NIQE), and<br>Inception Score |
| Wu *et al.*, 2020 | DCGAN | Learning rate=0.0002<br>Optimizer=Adam<br>Momentum=0.5<br>Batch size=32<br>Epochs=2,048 | GoogLeNet architecture achieved the highest identification accuracy of 94.33% on the test dataset;<br>The generated images from the DCGAN were qualitatively evaluated by plant experts, who confirmed that the synthetic images closely resembled real tomato leaf images |
| Zhang *et al.*, 2023 | GAN | Learning rate=0.0001<br>Batch Size=16<br>Epochs=1000<br>Activation Function=LeakyReLU<br>Regularization=L2<br>Dropout=0.5 (50%)<br>Optimizer=Adam<br>Decay rate=1e-8 | *Accuracy*=87%<br>Loss=0.2945<br>Precision=92.19% |
| Zhao *et al.*, 2022 | DoubleGAN (made up of WGAN and SRGAN) | Loss function optimizer=Wasserstein distance | Classification accuracy for the augmented dataset reaches 99.80%<br>Disease classification accuracy reached 99.53%. |

(*Contd...*)

Table 2: (*Continued*)

| Reference | Model | Training parameters | Performance metrics results |
|---|---|---|---|
| Zilvan *et al.*, 2019 | DC-VAE | Training the feature extractor involved varying the number of epochs at 50, 100, 150, and 200 The classifier was trained for a fixed 100 epochs. The models were trained independently for each plant type, and a Fully Connected Network (FCN) classifier was used with the same architecture and settings across all systems. | The DC-VAE achieved the following ACCURACIES: *For corn*: Natural Condition: 86.09%, Salt Noise: 86.32%, Pepper Noise: 87.97%, Salt & Pepper Noise: 87.26% *For potato*: Natural Condition: 86.77%, Salt Noise: 89.22%, Pepper Noise: 88.24%, Salt & Pepper Noise: 88.24% |

Marrewijk *et al.* (2022) focused on leaf scab on apple trees. This diversity spans multiple plant species and disease types, including fungal, bacterial, viral, and pest-related diseases. The relationship between dataset size and model performance varies across studies. Alshammari *et al.* (2024) used 3,505 images, achieving high accuracy (ResNet50=86.59%, VGG19=84.28%) and precision (ResNet50=85.17%, VGG19=85.47%). Bi & Hu (2020), despite a smaller training set of 873 images, saw significant improvements in model performance across experiments, with F1-Scores ranging from 0.46 to 0.81. Förster *et al.* (2019) utilized 5,250 image patches, achieving praiseworthy visual evaluation results. Krishnakumar and Balasubrahmanyan (2024) employed a large dataset of 54,303 images, resulting in robust denoising metrics (RMSE=0.32702, PSNR=58.7179). Lopes *et al.* (2023) used 700 plots, with the PlantPlotGAN model outperforming others in evaluation metrics. Van Marrewijk *et al.* (2022) used the Plant Pathology 2020 and 2021 datasets, showing varying performance with F1-Scores ranging from 0.55 to 0.69. Generally, larger datasets correlate with better model performance, though data augmentation and preprocessing techniques also play crucial roles. The robustness of model training parameters significantly impacts performance (Table 2). Bi and Hu (2020) employed extensive training parameters, including 700 epochs, a learning rate of 0.0001, and a batch size of 100, along with overfitting mitigation techniques, resulting in improved F1-Scores across experiments. Förster *et al.* (2019) used CycleGAN with adversarial and cycle-consistent loss functions, achieving high-quality image generation. Krishnakumar and Balasubrahmanyan (2024) implemented Hybrid Fourier Filter Denoising and EigenGAN, optimizing learning rate, batch size, and epochs for effective denoising. Lopes *et al.* (2023) applied multiple GAN models with specific optimizer settings (Adam, learning rate=2e-4, β1=0.5, β2=0.99), leading to superior performance in evaluation metrics. Van Marrewijk *et al.* (2022) utilized CycleGAN with detailed training parameters, achieving varying precision, recall, and F1-Scores across different datasets. Robust training parameters, including learning rates, batch sizes, epochs, and loss functions, are critical for enhancing model performance and achieving reliable results.

## CONCLUSION

This study highlights the vital role of collaboration and citation metrics in assessing the impact of researchers in deep learning for plant disease diagnosis. Key findings reveal significant contributions from prominent authors, underscoring the importance of addressing critical issues through innovative research. Future investigations should build on these insights to explore methodologies that enhance academic influence and practical advancements in the field.

## DATA AVAILABILITY

The dataset used for this study is available at https://github.com/Enowtakang/Review2024.

## REFERENCES

Abrar, S., & Samad, M. D. (2022). Perturbation of deep autoencoder weights for model compression and classification of tabular data. *Neural Networks, 156*, 160-169. https://doi.org/10.1016/j.neunet.2022.09.020

Ahmad, A., Saraswat, D., & El Gamal, A. (2023). A survey on using deep learning techniques for plant disease diagnosis and recommendations for development of appropriate tools. *Smart Agricultural Technology, 3*, 100083. https://doi.org/10.1016/j.atech.2022.100083

Ahmed, I., & Yadav, P. K. (2023). A systematic analysis of machine learning and deep learning based approaches for identifying and diagnosing plant diseases. *Sustainable Operations and Computers, 4*, 96-104. https://doi.org/10.1016/j.susoc.2023.03.001

Ali, T., Rehman, S. U., Ali, S., Mahmood, K., Obregon, S. A., Iglesias, R. C., Khurshaid, T., & Ashraf, I. (2024). Smart agriculture: Utilizing machine learning and deep learning for drought stress identification in crops. *Scientific Reports, 14*, 30062. https://doi.org/10.1038/s41598-024-74127-8

Alimisis, P., Mademlis, I., Radoglou-Grammatikis, P., Sarigiannidis, P., & Papadopoulos, G. T. (2024). Advances in Diffusion Models for Image Data Augmentation: A Review of Methods, Models, Evaluation Metrics and Future Research Directions. *arXiv*, arXiv:2407.04103. https://doi.org/10.48550/arXiv.2407.04103

Alshammari, K., Alshammari, R., Alshammari, A., & Alkhudaydi, T. (2024). An improved pear disease classification approach using cycle generative adversarial network. *Scientific Reports, 14*, 6680. https://doi.org/10.1038/s41598-024-57143-6

Altalak, M., Ammad uddin, M., Alajmi, A., & Rizg, A. (2022). Smart Agriculture Applications Using Deep Learning Technologies: A Survey. *Applied Sciences, 12*(12), 5919. https://doi.org/10.3390/app12125919

Alzahrani, M. S., & Alsaade, F. W. (2023). Transform and Deep Learning Algorithms for the Early Detection and Recognition of Tomato Leaf Disease. *Agronomy, 13*(5), 1184. https://doi.org/10.3390/agronomy13051184

Amara, J., König-Ries, B., & Samuel, S. (2024). Explainability of Deep Learning-Based Plant Disease Classifiers Through Automated Concept Identification. *arXiv*, arXiv:2412.07408. https://doi.

org/10.48550/arXiv.2412.07408
Amyar, A., Ruan, S., Vera, P., Decazes, P., & Modzelewski, R. (2020). RADIOGAN: Deep Convolutional Conditional Generative Adversarial Network To Generate PET Images. *arXiv*, arXiv:2003.08663. https://doi.org/10.48550/arXiv.2003.08663
An, T., & Joo, C. (2023). CycleGANAS: Differentiable Neural Architecture Search for CycleGAN. *arXiv*, arXiv:2311.07162. https://doi.org/10.48550/arXiv.2311.07162
Andrew, J., Eunice, J., Popescu, D. E., Chowdary, M. K., & Hemanth, J. (2022). Deep Learning-Based Leaf Disease Detection in Crops Using Images for Agricultural Applications. *Agronomy, 12*(10), 2395. https://doi.org/10.3390/agronomy12102395
Antoniou, A., Storkey, A., & Edwards, H. (2018). Data Augmentation Generative Adversarial Networks. *arXiv*, arXiv:1711.04340. https://doi.org/10.48550/arXiv.1711.04340
Aslan, S., Güdükbay, U., Töreyin, B. U., & Çetin, A. E. (2019). Deep Convolutional Generative Adversarial Networks Based Flame Detection in Video. *arXiv*, arXiv:1902.01824. https://doi.org/10.48550/arXiv.1902.01824
Baghel, N., Dubey, S. R., & Singh, S. K. (2023). SRTransGAN: Image Super-Resolution using Transformer based Generative Adversarial Network. *arXiv*, arXiv:2312.01999. https://doi.org/10.48550/arXiv.2312.01999
Baig, Y., Ma, H. R., Xu, H., & You, L. (2023). Autoencoder neural networks enable low dimensional structure analyses of microbial growth dynamics. *Nature Communications, 14*, 7937. https://doi.org/10.1038/s41467-023-43455-0
Bandyopadhyay, S., Dion, C., Libon, D. J., Price, C., Tighe, P., & Rashidi, P. (2022). Variational autoencoder provides proof of concept that compressing CDT to extremely low-dimensional space retains its ability of distinguishing dementia. *Scientific Reports, 12*, 7992. https://doi.org/10.1038/s41598-022-12024-8
Bank, D., Koenigstein, N., & Giryes, R. (2021). Autoencoders. *arXiv*, arXiv:2003.05991. https://doi.org/10.48550/arXiv.2003.05991
Benfenati, A., Causin, P., Oberti, R., & Stefanello, G. (2023). Unsupervised deep learning techniques for automatic detection of plant diseases: Reducing the need of manual labelling of plant images. *Journal of Mathematics in Industry, 13*, 5. https://doi.org/10.1186/s13362-023-00133-6
Berahmand, K., Daneshfar, F., Salehi, E. S., Li, Y., & Xu, Y. (2024). Autoencoders and their applications in machine learning: A survey. *Artificial Intelligence Review, 57*, 28. https://doi.org/10.1007/s10462-023-10662-6
Bertrand, J. H., Gargano, J. P., Mombaerts, L., & Taws, J. (2024). Autoencoder-based General Purpose Representation Learning for Customer Embedding. *arXiv*, arXiv:2402.18164. https://doi.org/10.48550/arXiv.2402.18164
Bhuvana, J., & Mirnalinee, T. T. (2021). An approach to Plant Disease Detection using Deep Learning Techniques. *Iteckne, 18*(2), 161-169. https://doi.org/10.15332/iteckne.v18i2.2615
Bi, L., & Hu, G. (2020). Improving Image-Based Plant Disease Classification With Generative Adversarial Network Under Limited Training Set. *Frontiers in Plant Science, 11*, 583438. https://doi.org/10.3389/fpls.2020.583438
Blarr, J., Klinder, S., Liebig, W. V., Inal, K., Kärger, L., & Weidenmann, K. A. (2024). Deep convolutional generative adversarial network for generation of computed tomography images of discontinuously carbon fiber reinforced polymer microstructures. *Scientific Reports, 14*, 9641. https://doi.org/10.1038/s41598-024-59252-8
Bolluyt, E. D., & Comaniciu, C. (2019). Collapse Resistant Deep Convolutional GAN for Multi-Object Image Generation. *arXiv*, arXiv:1911.02996. https://doi.org/10.48550/arXiv.1911.02996
Boulahbal, H. eddine, Voicila, A., & Comport, A. (2022). Are conditional GANs explicitly conditional? *arXiv*, arXiv:2106.15011. https://doi.org/10.48550/arXiv.2106.15011
Bourlard, H., & Kabil, S. H. (2022). Autoencoders reloaded. *Biological Cybernetics, 116*, 389-406. https://doi.org/10.1007/s00422-022-00937-6
Bunker, J., Girolami, M., Lambley, H., Stuart, A. M., & Sullivan, T. J. (2024). Autoencoders in Function Space. *arXiv*, arXiv:2408.01362. https://doi.org/10.48550/arXiv.2408.01362
Cakmak, A. S., Thigpen, N., Honke, G., Alday, E. P., Rad, A. B., Adaimi, R., Chang, C. J., Li, Q., Gupta, P., Neylan, T., McLean, S. A., & Clifford, G. D. (2020). Using Convolutional Variational Autoencoders to Predict Post-Trauma Health Outcomes from Actigraphy Data. *arXiv*, arXiv:2011.07406. https://doi.org/10.48550/arXiv.2011.07406
Casale, F. P., Dalca, A. V., Saglietti, L., Listgarten, J., & Fusi, N. (2018). Gaussian Process Prior Variational Autoencoders. *arXiv*, arXiv:1810.11738. https://doi.org/10.48550/arXiv.1810.11738
Cemgil, A. T., Ghaisas, S., Dvijotham, K., Gowal, S., & Kohli, P. (2020). Autoencoding Variational Autoencoder. *arXiv*, arXiv:2012.03715. https://doi.org/10.48550/arXiv.2012.03715
Chandrakala, S., & Vishnika, V. S. (2024). Denoising Convolutional Autoencoder Based Approach for Disordered Speech Recognition. *International Journal on Artificial Intelligence Tools, 33*(1), 2350058. https://doi.org/10.1142/S0218213023500586
Che, Q.-H., Le, D.-T., & Nguyen, V.-T. (2024). Enhanced Generative Data Augmentation for Semantic Segmentation via Stronger Guidance. *arXiv*, arXiv:2409.06002. https://doi.org/10.48550/arXiv.2409.06002
Chen, C., Liu, W., Tan, X., & Wong, K.-Y. K. (2023). Semi-supervised Cycle-GAN for face photo-sketch translation in the wild. *arXiv*, arXiv:2307.10281. https://doi.org/10.48550/arXiv.2307.10281
Chen, J., & Shi, X. (2019). Sparse Convolutional Denoising Autoencoders for Genotype Imputation. *Genes, 10*(9), 652. https://doi.org/10.3390/genes10090652
Chen, R., Qi, H., Liang, Y., & Yang, M. (2022). Identification of plant leaf diseases by deep learning based on channel attention and channel pruning. *Frontiers in Plant Science, 13*, 1023515. https://doi.org/10.3389/fpls.2022.1023515
Chen, S., & Guo, W. (2023). Auto-Encoders in Deep Learning—A Review with New Perspectives. *Mathematics, 11*(8), 1777. https://doi.org/10.3390/math11081777
Chen, Y., Yan, Z., & Zhu, Y. (2024). A Unified Framework for Generative Data Augmentation: A Comprehensive Survey. *arXiv*, arXiv:2310.00277. https://doi.org/10.48550/arXiv.2310.00277
Chen, Z., & Tong, Y. (2017). Face Super-Resolution Through Wasserstein GANs. *arXiv*, arXiv:1705.02438. https://doi.org/10.48550/arXiv.1705.02438
Cheng, L., Guan, P., Taherkordi, A., Liu, L., & Lan, D. (2024). Variational autoencoder-based neural network model compression. *arXiv*, arXiv:2408.14513. https://doi.org/10.48550/arXiv.2408.14513
Chiba, N., Suzuki, Y., Taniai, T., Igarashi, R., Ushiku, Y., Saito, K., & Ono, K. (2023). Neural structure fields with application to crystal structure autoencoders. *Communications Materials, 4*, 106. https://doi.org/10.1038/s43246-023-00432-w
Choi, J., Kim, Y., Kim, K.-H., Jung, S.-H., & Cho, I. (2023). PCT-CycleGAN: Paired Complementary Temporal Cycle-Consistent Adversarial Networks for Radar-Based Precipitation Nowcasting. arXiv, arXiv:2211.15046. https://doi.org/10.48550/arXiv.2211.15046
Creswell, A., Arulkumaran, K., & Bharath, A. A. (2017). On denoising autoencoders trained to minimise binary cross-entropy. *arXiv*, arXiv:1708.08487. https://doi.org/10.48550/arXiv.1708.08487
Cunningham, H., Ewart, A., Riggs, L., Huben, R., & Sharkey, L. (2023). Sparse Autoencoders Find Highly Interpretable Features in Language Models. *arXiv*, arXiv:2309.08600. https://doi.org/10.48550/arXiv.2309.08600
Curtó, J. D., Zarza, I. C., Torre, F. de la, King, I., & Lyu, M. R. (2020). High-resolution Deep Convolutional Generative Adversarial Networks. *arXiv*, arXiv:1711.06491. https://doi.org/10.48550/arXiv.1711.06491
D'Angelo, G., & Palmieri, F. (2021). A stacked autoencoder-based convolutional and recurrent deep neural network for detecting cyberattacks in interconnected power control systems. *International Journal of Intelligent Systems, 36*, 7080-7102. https://doi.org/10.1002/int.22581
Dai, D., Xia, P., Zhu, Z., & Che, H. (2023). MTDL-EPDCLD: A Multi-Task Deep-Learning-Based System for Enhanced Precision Detection and Diagnosis of Corn Leaf Diseases. *Plants, 12*(13), 2433. https://doi.org/10.3390/plants12132433
DeVries, T., Romero, A., Pineda, L., Taylor, G. W., & Drozdzal, M. (2019). On the Evaluation of Conditional GANs. *arXiv*, arXiv:1907.08175. https://doi.org/10.48550/arXiv.1907.08175
Dhaka, V. S., Kundu, N., Rani, G., Zumpano, E., & Vocaturo, E. (2023a). Role of Internet of Things and Deep Learning Techniques in Plant Disease Detection and Classification: A Focused Review. *Sensors, 23*(18), 7877. https://doi.org/10.3390/s23187877
Doersch, C. (2021). Tutorial on Variational Autoencoders. *arXiv*, arXiv:1606.05908. https://doi.org/10.48550/arXiv.1606.05908
Förster, A., Behley, J., Behmann, J., & Roscher, R. (2019). Hyperspectral Plant Disease Forecasting Using Generative Adversarial Networks.

*IGARSS 2019 - 2019 IEEE International Geoscience and Remote Sensing Symposium*, 1793-1796. https://doi.org/10.1109/IGARSS.2019.8898749

Frizza, T., Dansereau, D. G., Seresht, N. M., & Bewley, M. (2022). Semantically Accurate Super-Resolution Generative Adversarial Networks. *arXiv*, arXiv:2205.08659. https://doi.org/10.48550/arXiv.2205.08659

Fu, Y., Chen, C., Qiao, Y., & Yu, Y. (2024). DreamDA: Generative Data Augmentation with Diffusion Models. *arXiv*, arXiv:2403.12803. https://doi.org/10.48550/arXiv.2403.12803

Gao, L., Tour, T. D. la, Tillman, H., Goh, G., Troll, R., Radford, A., Sutskever, I., Leike, J., & Wu, J. (2024). Scaling and evaluating sparse autoencoders. *arXiv*, arXiv:2406.04093. https://doi.org/10.48550/arXiv.2406.04093

Ghafar, A., Chen, C., Shah, S. A. A., Ur Rehman, Z., & Rahman, G. (2024). Visualizing Plant Disease Distribution and Evaluating Model Performance for Deep Learning Classification with YOLOv8. *Pathogens, 13*(12), 1032. https://doi.org/10.3390/pathogens13121032

Ghosh, P., Sajjadi, M. S. M., Vergari, A., Black, M., & Schölkopf, B. (2020). From Variational to Deterministic Autoencoders. *arXiv*, arXiv:1903.12436. https://doi.org/10.48550/arXiv.1903.12436

Girin, L., Leglaive, S., Bie, X., Diard, J., Hueber, T., & Alameda-Pineda, X. (2022). Dynamical Variational Autoencoders: A Comprehensive Review. *arXiv*, arXiv:2008.12595. https://doi.org/10.48550/arXiv.2008.12595

Giuliano, A., Gadsden, S. A., Hilal, W., & Yawney, J. (2024). Convolutional variational autoencoders for secure lossy image compression in remote sensing. *arXiv*, arXiv:2404.03696. https://doi.org/10.48550/arXiv.2404.03696

Gomari, D. P., Schweickart, A., Cerchietti, L., Paietta, E., Fernandez, H., Al-Amin, H., Suhre, K., & Krumsiek, J. (2022). Variational autoencoders learn transferrable representations of metabolomics data. *Communications Biology, 5*, 645. https://doi.org/10.1038/s42003-022-03579-3

Gong, Y., Zhan, Z., Jin, Q., Li, Y., Idelbayev, Y., Liu, X., Zharkov, A., Aberman, K., Tulyakov, S., Wang, Y., & Ren, J. (2024). E$^2$GAN: Efficient Training of Efficient GANs for Image-to-Image Translation. *arXiv*, arXiv:2401.06127. https://doi.org/10.48550/arXiv.2401.06127

González-Rodríguez, V. E., Izquierdo-Bueno, I., Cantoral, J. M., Carbú, M., & Garrido, C. (2024). Artificial Intelligence: A Promising Tool for Application in Phytopathology. *Horticulturae, 10*(3), 197. https://doi.org/10.3390/horticulturae10030197

Goodfellow, I. J., Pouget-Abadie, J., Mirza, M., Xu, B., Warde-Farley, D., Ozair, S., Courville, A., & Bengio, Y. (2014). Generative Adversarial Networks. *arXiv*, arXiv:1406.2661. https://doi.org/10.48550/arXiv.1406.2661

Gu, Y. H., Yin, H., Jin, D., Zheng, R., & Yoo, S. J. (2022). Improved Multi-Plant Disease Recognition Method Using Deep Convolutional Neural Networks in Six Diseases of Apples and Pears. *Agriculture, 12*(2), 300. https://doi.org/10.3390/agriculture12020300

Güemes, A., Vila, C. S., & Discetti, S. (2022). Super-resolution GANs of randomly-seeded fields. *arXiv*, arXiv:2202.11701. https://doi.org/10.48550/arXiv.2202.11701

Gulamali, F. F., Sawant, A. S., Kovatch, P., Glicksberg, B., Charney, A., Nadkarni, G. N., & Oermann, E. (2022). Autoencoders for sample size estimation for fully connected neural network classifiers. *NPJ Digital Medicine, 5*, 180. https://doi.org/10.1038/s41746-022-00728-0

Gulrajani, I., Ahmed, F., Arjovsky, M., Dumoulin, V., & Courville, A. (2017). Improved Training of Wasserstein GANs. *arXiv*, arXiv:1704.00028. https://doi.org/10.48550/arXiv.1704.00028

Guo, Y., Zhang, J., Yin, C., Hu, X., Zou, Y., Xue, Z., & Wang, W. (2020). Plant Disease Identification Based on Deep Learning Algorithm in Smart Farming. *Discrete Dynamics in Nature and Society, 2020*, 2479172. https://doi.org/10.1155/2020/2479172

Hasan, R. I., Yusuf, S. M., & Alzubaidi, L. (2020). Review of the State of the Art of Deep Learning for Plant Diseases: A Broad Analysis and Discussion. *Plants, 9*(10), 1302. https://doi.org/10.3390/plants9101302

He, Y., Zhang, Z., Zhu, J., Shen, Y., & Chen, Q. (2022). Interpreting Class Conditional GANs with Channel Awareness. *arXiv*, arXiv:2203.11173. https://doi.org/10.48550/arXiv.2203.11173

He, Z., Kan, M., & Shan, S. (2021). EigenGAN: Layer-Wise Eigen-Learning for GANs. *arXiv*, arXiv:2104.12476. https://doi.org/10.48550/arXiv.2104.12476

Hemalatha, S., & Jayachandran, J. J. B. (2024). A Multitask Learning-Based Vision Transformer for Plant Disease Localization and Classification. *International Journal of Computational Intelligence Systems, 17*(1), 188. https://doi.org/10.1007/s44196-024-00597-3

Hou, L., Cao, Q., Shen, H., Pan, S., Li, X., & Cheng, X. (2022). Conditional GANs with Auxiliary Discriminative Classifier. *arXiv*, arXiv:2107.10060. https://doi.org/10.48550/arXiv.2107.10060

Hu, Y., Chu, X., & Zhang, B. (2024). Masked Autoencoders Are Robust Neural Architecture Search Learners. *arXiv*, arXiv:2311.12086. https://doi.org/10.48550/arXiv.2311.12086

Huang, G., & Jafari, A. H. (2021). Enhanced balancing GAN: Minority-class image generation. *Neural Computing and Applications, 35*, 5145-5145. https://doi.org/10.1007/s00521-021-06163-8

Huang, T., Chakraborty, P., & Sharma, A. (2020). Deep convolutional generative adversarial networks for traffic data imputation encoding time series as images. *arXiv*, arXiv:2005.04188. https://doi.org/10.48550/arXiv.2005.04188

Huang, Y., & Omachi, S. (2023). Infrared Image Super-Resolution via GAN. *arXiv*, arXiv:2312.00689. https://doi.org/10.48550/arXiv.2312.00689

Hughes, D. P., & Salathe, M. (2016). An open access repository of images on plant health to enable the development of mobile disease diagnostics. *arXiv*, arXiv:1511.08060. https://doi.org/10.48550/arXiv.1511.08060

Iacono, P., & Khan, N. (2023). Structure Preserving Cycle-GAN for Unsupervised Medical Image Domain Adaptation. *arXiv*, arXiv:2304.09164. https://doi.org/10.48550/arXiv.2304.09164

Iakovenko, O., & Bondarenko, I. (2024). Convolutional Variational Autoencoders for Spectrogram Compression in Automatic Speech Recognition. *arXiv*, arXiv:2410.02560. https://doi.org/10.48550/arXiv.2410.02560

Isola, P., Zhu, J.-Y., Zhou, T., & Efros, A. A. (2018). Image-to-Image Translation with Conditional Adversarial Networks. *arXiv*, arXiv:1611.07004. https://doi.org/10.48550/arXiv.1611.07004

Jafar, A., Bibi, N., Naqvi, R. A., Sadeghi-Niaraki, A., & Jeong, D. (2024). Revolutionizing agriculture with artificial intelligence: Plant disease detection methods, applications, and their limitations. *Frontiers in Plant Science, 15*, 1356260. https://doi.org/10.3389/fpls.2024.1356260

Jolicoeur-Martineau, A., & Mitliagkas, I. (2020). Gradient penalty from a maximum margin perspective. *arXiv*, arXiv:1910.06922. https://doi.org/10.48550/arXiv.1910.06922

Jung, M., Song, J. S., Shin, A.-Y., Choi, B., Go, S., Kwon, S.-Y., Park, J., Park, S. G., & Kim, Y.-M. (2023). Construction of deep learning-based disease detection model in plants. *Scientific Reports, 13*, 7331. https://doi.org/10.1038/s41598-023-34549-2

Kang, M., Zhang, R., Barnes, C., Paris, S., Kwak, S., Park, J., Shechtman, E., Zhu, J.-Y., & Park, T. (2024). Distilling Diffusion Models into Conditional GANs. *arXiv*, arXiv:2405.05967. https://doi.org/10.48550/arXiv.2405.05967

Kas, M., Chahi, A., Kajo, I., & Ruichek, Y. (2024). EigenGAN: An SVD subspace-based learning for image generation using Conditional GAN. *Knowledge-Based Systems, 293*, 111691. https://doi.org/10.1016/j.knosys.2024.111691

Kebaili, A., Lapuyade-Lahorgue, J., & Ruan, S. (2023). Deep Learning Approaches for Data Augmentation in Medical Imaging: A Review. *arXiv*, arXiv:2307.13125. https://doi.org/10.48550/arXiv.2307.13125

Kim, C., Park, S., & Hwang, H. J. (2018). Local Stability and Performance of Simple Gradient Penalty mu-Wasserstein GAN. *arXiv*, arXiv:1810.02528. https://doi.org/10.48550/arXiv.1810.02528

Kinakh, V., Drozdova, M., Quétant, G., Golling, T., & Voloshynovskiy, S. (2021). Information-theoretic stochastic contrastive conditional GAN: InfoSCC-GAN. *arXiv*, arXiv:2112.09653. https://doi.org/10.48550/arXiv.2112.09653

Kingma, D. P., & Welling, M. (2019). An Introduction to Variational Autoencoders. *arXiv*, arXiv:1906.02691. https://doi.org/10.48550/arXiv.1906.02691

Kingma, D. P., & Welling, M. (2022). Auto-Encoding Variational Bayes. *arXiv*, arXiv:1312.6114. https://doi.org/10.48550/arXiv.1312.6114

Kipf, T. N., & Welling, M. (2016). Variational Graph Auto-Encoders. *arXiv*, arXiv:1611.07308. https://doi.org/10.48550/arXiv.1611.07308

Krishna, P. A., Padhy, N., & Patnaik, A. (2024). Applying Machine Learning for Sustainable Farm Management: Integrating Crop Recommendations and Disease Identification. *Engineering Proceedings, 67*(1), 7073. https://doi.org/10.3390/engproc2024067073

Krishnakumar, D. P., & Balasubrahmanyan, K. (2024). An Improved EigenGAN-Based Method for Data Augmentation for Plant Disease

Classification. *International Information and Engineering Technology Association, 38*(1), 237-242. https://doi.org/10.18280/ria.380124
Kuznedelev, D., Startsev, V., Shlenskii, D., & Kastryulin, S. (2024). Does Diffusion Beat GAN in Image Super Resolution? *arXiv*, arXiv:2405.17261. https://doi.org/10.48550/arXiv.2405.17261
Kwon, D., Kim, Y., Montúfar, G., & Yang, I. (2021). Training Wasserstein GANs without gradient penalties. *arXiv*, arXiv:2110.14150. https://doi.org/10.48550/arXiv.2110.14150
Lebrini, Y., & Ayerdi Gotor, A. (2024). Crops Disease Detection, from Leaves to Field: What We Can Expect from Artificial Intelligence. *Agronomy, 14*(11), 2719. https://doi.org/10.3390/agronomy14112719
Ledig, C., Theis, L., Huszar, F., Caballero, J., Cunningham, A., Acosta, A., Aitken, A., Tejani, A., Totz, J., Wang, Z., & Shi, W. (2016). Photo-Realistic Single Image Super-Resolution Using a Generative Adversarial Network. *arXiv*, arXiv:1609.04802. https://doi.org/10.48550/arXiv.1609.04802
Lee, D., Choi, S., & Kim, H.-J. (2018). Performance evaluation of image denoising developed using convolutional denoising autoencoders in chest radiography. *Nuclear Instruments and Methods in Physics Research Section A: Accelerators, Spectrometers, Detectors and Associated Equipment, 884*, 97-104. https://doi.org/10.1016/j.nima.2017.12.050
Lee, Y. (2023). A Geometric Perspective on Autoencoders. *arXiv*, arXiv:2309.08247. https://doi.org/10.48550/arXiv.2309.08247
Lei, H., & Yang, Y. (2021). CDAE: A Cascade of Denoising Autoencoders for Noise Reduction in the Clustering of Single-Particle Cryo-EM Images. *Frontiers in Genetics, 11*, 627746. https://doi.org/10.3389/fgene.2020.627746
Lingenberg, T., Reuter, M., Sudhakaran, G., Gojny, D., Roth, S., & Schaub-Meyer, S. (2024). DIAGen: Diverse Image Augmentation with Generative Models. *arXiv*, arXiv:2408.14584. https://doi.org/10.48550/arXiv.2408.14584
Liu, B., Wei, S., Zhang, F., Guo, N., Fan, H., & Yao, W. (2024). Tomato leaf disease recognition based on multi-task distillation learning. *Frontiers in Plant Science, 14*, 1330527. https://doi.org/10.3389/fpls.2023.1330527
Lopes, F. A., Sagan, V., & Esposito, F. (2023). PlantPlotGAN: A Physics-Informed Generative Adversarial Network for Plant Disease Prediction. *arXiv*, arXiv:2310.18268. https://doi.org/10.48550/arXiv.2310.18268
Makhzani, A., Shlens, J., Jaitly, N., Goodfellow, I., & Frey, B. (2016). Adversarial Autoencoders. *arXiv*, arXiv:1511.05644. https://doi.org/10.48550/arXiv.1511.05644
Manduchi, L., Vandenhirtz, M., Ryser, A., & Vogt, J. (2023). Tree Variational Autoencoders. *arXiv*, arXiv:2306.08984. https://doi.org/10.48550/arXiv.2306.08984
Martino, G., Moroni, D., & Martinelli, M. (2023). Are We Using Autoencoders in a Wrong Way? *arXiv*, arXiv:2309.01532. https://doi.org/10.48550/arXiv.2309.01532
Merkelbach, K., Schaper, S., Diedrich, C., Fritsch, S. J., & Schuppert, A. (2023). Novel architecture for gated recurrent unit autoencoder trained on time series from electronic health records enables detection of ICU patient subgroups. *Scientific Reports, 13*, 4053. https://doi.org/10.1038/s41598-023-30986-1
Michelucci, U. (2022). An Introduction to Autoencoders. *arXiv*, arXiv:2201.03898. https://doi.org/10.48550/arXiv.2201.03898
Milne, T., & Nachman, A. (2022). Wasserstein GANs with Gradient Penalty Compute Congested Transport. *arXiv*, arXiv:2109.00528. https://doi.org/10.48550/arXiv.2109.00528
Min, B., Kim, T., Shin, D., & Shin, D. (2023). Data Augmentation Method for Plant Leaf Disease Recognition. *Applied Sciences, 13*(3), 1465. https://doi.org/10.3390/app13031465
Miranda, M., Zabawa, L., Kicherer, A., Strothmann, L., Rascher, U., & Roscher, R. (2022). Detection of Anomalous Grapevine Berries Using Variational Autoencoders. *Frontiers in Plant Science, 13*, 729097. https://doi.org/10.3389/fpls.2022.729097
Mirza, B., Haroon, D., Khan, B., Padhani, A., & Syed, T. Q. (2021). Deep Generative Models to Counter Class Imbalance: A Model-Metric Mapping With Proportion Calibration Methodology. *IEEE Access, 9*, 55879-55897. https://doi.org/10.1109/ACCESS.2021.3071389
Mirza, M., & Osindero, S. (2014). Conditional Generative Adversarial Nets. *arXiv*, arXiv:1411.1784. https://doi.org/10.48550/arXiv.1411.1784
Mishra, A., & Pathak, T. (2021). Deep Convolutional Generative Modeling for Artificial Microstructure Development of Aluminum-Silicon Alloy. *arXiv*, arXiv:2109.06635. https://doi.org/10.48550/arXiv.2109.06635
Mohanty, S. P., Hughes, D. P., & Salathé, M. (2016). Using Deep Learning for Image-Based Plant Disease Detection. *Frontiers in Plant Science, 7*, 1419. https://doi.org/10.3389/fpls.2016.01419
Moyes, A., Gault, R., Zhang, K., Ming, J., Crookes, D., & Wang, J. (2023). Multi-channel auto-encoders for learning domain invariant representations enabling superior classification of histopathology images. *Medical Image Analysis, 83*, 102640. https://doi.org/10.1016/j.media.2022.102640
Mustofa, S., Munna, M. M. H., Emon, Y. R., Rabbany, G., & Ahad, M. T. (2023). A comprehensive review on Plant Leaf Disease detection using Deep learning. *arXiv*, arXiv:2308.14087. https://doi.org/10.48550/arXiv.2308.14087
Myers, J., Najafian, K., Maleki, F., & Ovens, K. (2024). Modified CycleGAN for the synthesization of samples for wheat head segmentation. *arXiv*, arXiv:2402.15135. https://doi.org/10.48550/arXiv.2402.15135
Nagaraj, G., Sungeetha, D., Tiwari, M., Ahuja, V., Varma, A. K., & Agarwal, P. (2024). Advancements in Plant Pests Detection: Leveraging Convolutional Neural Networks for Smart Agriculture. *Engineering Proceedings, 59*(1), 201. https://doi.org/10.3390/engproc2023059201
Nanavaty, A., Sharma, R., Pandita, B., Goyal, O., Rallapalli, S., Mandal, M., Singh, V. K., Narang, P., & Chamola, V. (2024). Integrating deep learning for visual question answering in Agricultural Disease Diagnostics: Case Study of Wheat Rust. *Scientific Reports, 14*, 28203. https://doi.org/10.1038/s41598-024-79793-2
Natarajan, S., Chakrabarti, P., & Margala, M. (2024). Robust diagnosis and meta visualizations of plant diseases through deep neural architecture with explainable AI. *Scientific Reports, 14*, 13695. https://doi.org/10.1038/s41598-024-64601-8
Ngugi, H. N., Akinyelu, A. A., & Ezugwu, A. E. (2024). Machine Learning and Deep Learning for Crop Disease Diagnosis: Performance Analysis and Review. *Agronomy, 14*(12), 3001. https://doi.org/10.3390/agronomy14123001
Nguyen, C., Sagan, V., Maimaitiyiming, M., Maimaitijiang, M., Bhadra, S., & Kwasniewski, M. T. (2021). Early Detection of Plant Viral Disease Using Hyperspectral Imaging and Deep Learning. *Sensors, 21*(3), 742. https://doi.org/10.3390/s21030742
Nigam, S., & Jain, R. (2020). Plant disease identification using Deep Learning: A review. *The Indian Journal of Agricultural Sciences, 90*(2), 249-257. https://doi.org/10.56093/ijas.v90i2.98996
Omer, S. M., Ghafoor, K. Z., & Askar, S. K. (2023). Plant Disease Diagnosing Based on Deep Learning Techniques. *Aro-The Scientific Journal of Koya University, 11*(1), 38-47. https://doi.org/10.14500/aro.11080
Pacal, I., Kunduracioglu, I., Alma, M. H., Deveci, M., Kadry, S., Nedoma, J., Slany, V., & Martinek, R. (2024). A systematic review of deep learning techniques for plant diseases. *Artificial Intelligence Review, 57*, 304. https://doi.org/10.1007/s10462-024-10944-7
Pan, T., Pedrycz, W., Yang, J., & Wang, J. (2024). An improved generative adversarial network to oversample imbalanced datasets. *Engineering Applications of Artificial Intelligence, 132*, 107934. https://doi.org/10.1016/j.engappai.2024.107934
Papadopoulos, D., & Karalis, V. D. (2023). Variational Autoencoders for Data Augmentation in Clinical Studies. *Applied Sciences, 13*(15), 8793. https://doi.org/10.3390/app13158793
Prajapati, H. B., Shah, J. P., & Dabhi, V. K. (2017). Detection and classification of rice plant diseases. *Intelligent Decision Technologies, 11*(3), 357-373. https://doi.org/10.3233/IDT-170301
Prakash, M., Krull, A., & Jug, F. (2021). Fully Unsupervised Diversity Denoising with Convolutional Variational Autoencoders. *arXiv*, arXiv:2006.06072. https://doi.org/10.48550/arXiv.2006.06072
Qin, Y., Li, Z., Xie, S., Zhao, H., & Wang, Q. (2024). An Efficient Convolutional Denoising Autoencoder-Based BDS NLOS Detection Method in Urban Forest Environments. *Sensors, 24*(6), 1959. https://doi.org/10.3390/s24061959
Radford, A., Metz, L., & Chintala, S. (2016). Unsupervised Representation Learning with Deep Convolutional Generative Adversarial Networks. *arXiv*, arXiv:1511.06434. https://doi.org/10.48550/arXiv.1511.06434
Radočaj, P., Radočaj, D., & Martinović, G. (2024). Image-Based Leaf Disease Recognition Using Transfer Deep Learning with a Novel Versatile Optimization Module. *Big Data and Cognitive Computing, 8*(6), 52. https://doi.org/10.3390/bdcc8060052
Rahat, F., Hossain, M. S., Ahmed, M. R., Jha, S. K., & Ewetz, R. (2024). Data Augmentation for Image Classification using Generative AI. *arXiv*, arXiv:2409.00547. https://doi.org/10.48550/arXiv.2409.00547
Rakhmatulin, I. (2022). Cycle-GAN for eye-tracking. *arXiv*, arXiv:2205.10556.

https://doi.org/10.48550/arXiv.2205.10556
Ramanjot, Mittal, U., Wadhawan, A., Singla, J., Jhanjhi, N. Z., Ghoniem, R. M., Ray, S. K., & Abdelmaboud, A. (2023). Plant Disease Detection and Classification: A Systematic Literature Review. *Sensors, 23*(10), 4769. https://doi.org/10.3390/s23104769
Rao, K., Harris, C., Irpan, A., Levine, S., Ibarz, J., & Khansari, M. (2020). RL-CycleGAN: Reinforcement Learning Aware Simulation-To-Real. *arXiv*, arXiv:2006.09001. https://doi.org/10.48550/arXiv.2006.09001
Rehana, H., Ibrahim, M., & Ali, M. H. (2023). Plant Disease Detection using Region-Based Convolutional Neural Network. *arXiv*, arXiv:2303.09063. https://doi.org/10.48550/arXiv.2303.09063
Rehm, F., Vallecorsa, S., Borras, K., & Krücker, D. (2021). Validation of Deep Convolutional Generative Adversarial Networks for High Energy Physics Calorimeter Simulations. *arXiv*, arXiv:2103.13698. https://doi.org/10.48550/arXiv.2103.13698
Rodríguez-Lira, D.-C., Córdova-Esparza, D.-M., Álvarez-Alvarado, J. M., Terven, J., Romero-González, J.-A., & Rodríguez-Reséndiz, J. (2024). Trends in Machine and Deep Learning Techniques for Plant Disease Identification: A Systematic Review. *Agriculture, 14*(12), 2188. https://doi.org/10.3390/agriculture14122188
Rudolph, M., Wandt, B., & Rosenhahn, B. (2019). Structuring Autoencoders. *arXiv*, arXiv:1908.02626. https://doi.org/10.48550/arXiv.1908.02626
Sagar, S., Javed, M., & Doermann, D. S. (2023). Leaf-Based Plant Disease Detection and Explainable AI. *arXiv*, arXiv:2404.16833. https://doi.org/10.48550/arXiv.2404.16833
Sajitha, P., Andrushia, A. D., Anand, N., & Naser, M. Z. (2024). A review on machine learning and deep learning image-based plant disease classification for industrial farming systems. *Journal of Industrial Information Integration, 38*, 100572. https://doi.org/10.1016/j.jii.2024.100572
Saleem, M. H., Khanchi, S., Potgieter, J., & Arif, K. M. (2020). Image-Based Plant Disease Identification by Deep Learning Meta-Architectures. *Plants, 9*(11), 1451. https://doi.org/10.3390/plants9111451
Saleem, M. H., Potgieter, J., & Arif, K. M. (2019). Plant Disease Detection and Classification by Deep Learning. *Plants, 8*(11), 468. https://doi.org/10.3390/plants8110468
Saleem, S., Sharif, M. I., Sharif, M. I., Sajid, M. Z., & Marinello, F. (2024). Comparison of Deep Learning Models for Multi-Crop Leaf Disease Detection with Enhanced Vegetative Feature Isolation and Definition of a New Hybrid Architecture. *Agronomy, 14*(10), 2230. https://doi.org/10.3390/agronomy14102230
Sandfort, V., Yan, K., Graffy, P. M., Pickhardt, P. J., & Summers, R. M. (2021). Use of Variational Autoencoders with Unsupervised Learning to Detect Incorrect Organ Segmentations at CT. *Radiology. Artificial Intelligence, 3*(4), e200218. https://doi.org/10.1148/ryai.2021200218
Shete, S., Srinivasan, S., & Gonsalves, T. A. (2020). TasselGAN: An Application of the Generative Adversarial Model for Creating Field-Based Maize Tassel Data. *Plant Phenomics, 2020*, 8309605. https://doi.org/10.34133/2020/8309605
Shi, C., Xu, T., Bergsma, W., & Li, L. (2021). Double Generative Adversarial Networks for Conditional Independence Testing. *arXiv*, arXiv:2006.02615. https://doi.org/10.48550/arXiv.2006.02615
Shoaib, M., Shah, B., El-Sappagh, S., Ali, A., Ullah, A., Alenezi, F., Gechev, T., Hussain, T., & Ali, F. (2023). An advanced deep learning models-based plant disease detection: A review of recent research. *Frontiers in Plant Science, 14*, 1158933. https://doi.org/10.3389/fpls.2023.1158933
Sim, B., Oh, G., Kim, J., Jung, C., & Ye, J. C. (2020). Optimal Transport driven CycleGAN for Unsupervised Learning in Inverse Problems. *arXiv*, arXiv:1909.12116. https://doi.org/10.48550/arXiv.1909.12116
Singh, A., & Ogunfunmi, T. (2021). An Overview of Variational Autoencoders for Source Separation, Finance, and Bio-Signal Applications. *Entropy, 24*(1), 55. https://doi.org/10.3390/e24010055
Skopek, O., Ganea, O.-E., & Bécigneul, G. (2020). Mixed-curvature Variational Autoencoders. *arXiv*, arXiv:1911.08411. https://doi.org/10.48550/arXiv.1911.08411
Song, J., & Ye, J. C. (2021). Federated CycleGAN for Privacy-Preserving Image-to-Image Translation. *arXiv*, arXiv:2106.09246. https://doi.org/10.48550/arXiv.2106.09246
Song, Z., Wang, D., Xiao, L., Zhu, Y., Cao, G., & Wang, Y. (2023). DaylilyNet: A Multi-Task Learning Method for Daylily Leaf Disease Detection. *Sensors, 23*(18), 7879. https://doi.org/10.3390/s23187879
Sun, R., Zhang, M., Yang, K., & Liu, J. (2020). Data Enhancement for Plant Disease Classification Using Generated Lesions. *Applied Sciences, 10*(2), 466. https://doi.org/10.3390/app10020466
Sun, X., Li, H., & Lee, W.-N. (2023). Constrained CycleGAN for Effective Generation of Ultrasound Sector Images of Improved Spatial Resolution. *arXiv*, arXiv:2309.00995. https://doi.org/10.48550/arXiv.2309.00995
Tadem, S. P. (2022). CycleGAN with three different unpaired datasets. *arXiv*, arXiv:2208.06526. https://doi.org/10.48550/arXiv.2208.06526
Thakur, P. S., Khanna, P., Sheorey, T., & Ojha, A. (2022). Explainable vision transformer enabled convolutional neural network for plant disease identification: PlantXViT. *arXiv*, arXiv:2207.07919. https://doi.org/10.48550/ArXiv.2207.07919
Tian, C., Zhang, X., Lin, J. C.-W., Zuo, W., Zhang, Y., & Lin, C.-W. (2022). Generative Adversarial Networks for Image Super-Resolution: A Survey. *arXiv*, arXiv:2204.13620. https://doi.org/10.48550/arXiv.2204.13620
Torbunov, D., Huang, Y., Tseng, H.-H., Yu, H., Huang, J., Yoo, S., Lin, M., Viren, B., & Ren, Y. (2023). UVCGAN v2: An Improved Cycle-Consistent GAN for Unpaired Image-to-Image Translation. *arXiv*, arXiv:2303.16280. https://doi.org/10.48550/arXiv.2303.16280
Torbunov, D., Huang, Y., Yu, H., Huang, J., Yoo, S., Lin, M., Viren, B., & Ren, Y. (2022). UVCGAN: UNet Vision Transformer cycle-consistent GAN for unpaired image-to-image translation. *arXiv*, arXiv:2203.02557. https://doi.org/10.48550/arXiv.2203.02557
van Marrewijk, B. M., Polder, G., & Kootstra, G. (2022). Investigation of the added value of CycleGAN on the plant pathology dataset. *IFAC-PapersOnLine, 55*(32), 89-94. https://doi.org/10.1016/j.ifacol.2022.11.120
Vankara, J., Nandini, S. S., Muddada, M. K., Kuppili, N. S. C., & Naidu, K. S. (2023). Plant Disease Prognosis Using Spatial-Exploitation-Based Deep-Learning Models. *Engineering Proceedings, 59*(1), 137. https://doi.org/10.3390/engproc2023059137
Vasudevan, N., & Karthick, T. (2023). A Hybrid Approach for Plant Disease Detection Using E-GAN and CapsNet. *Computer Systems Science and Engineering, 46*(1), 337-356. https://doi.org/10.32604/csse.2023.034242
Venkataraman, P. (2022). Image Denoising Using Convolutional Autoencoder. *arXiv*, arXiv:2207.11771. https://doi.org/10.48550/arXiv.2207.11771
Venu, S. K. (2020). Evaluation of Deep Convolutional Generative Adversarial Networks for data augmentation of chest X-ray images. *arXiv*, arXiv:2009.01181. https://doi.org/10.48550/arXiv.2009.01181
Wan, Q., Guo, W., & Wang, Y. (2024). SGBGAN: Minority class image generation for class-imbalanced datasets. *Machine Vision and Applications, 35*, 22. https://doi.org/10.1007/s00138-023-01506-y
Wang, R., Schmedding, S., & Huber, M. F. (2023a). Improving the Effectiveness of Deep Generative Data. *arXiv*, arXiv:2311.03959. https://doi.org/10.48550/arXiv.2311.03959
Wang, T., & Lin, Y. (2024). CycleGAN with Better Cycles. *arXiv*, arXiv:2408.15374. https://doi.org/10.48550/arXiv.2408.15374
Wang, X., & Cao, W. (2023). GACN: Generative Adversarial Classified Network for Balancing Plant Disease Dataset and Plant Disease Recognition. *Sensors, 23*(15), 6844. https://doi.org/10.3390/s23156844
Wang, X., Sun, L., Chehri, A., & Song, Y. (2023b). A Review of GAN-Based Super-Resolution Reconstruction for Optical Remote Sensing Images. *Remote Sensing, 15*(20), 5062. https://doi.org/10.3390/rs15205062
Wang, X., Yu, K., Wu, S., Gu, J., Liu, Y., Dong, C., Loy, C. C., Qiao, Y., & Tang, X. (2018). ESRGAN: Enhanced Super-Resolution Generative Adversarial Networks. *arXiv*, arXiv:1809.00219. https://doi.org/10.48550/arXiv.1809.00219
Wu, Q., Chen, Y., & Meng, J. (2020). DCGAN-Based Data Augmentation for Tomato Leaf Disease Identification. *IEEE Access, 8*, 98716-98728. https://doi.org/10.1109/ACCESS.2020.2997001
Yang, W., Yuan, Y., Zhang, D., Zheng, L., & Nie, F. (2024). An Effective Image Classification Method for Plant Diseases with Improved Channel Attention Mechanism aECAnet Based on Deep Learning. *Symmetry, 16*(4), 451. https://doi.org/10.3390/sym16040451
Yuan, Y., Sun, J., & Zhang, Q. (2024). An Enhanced Deep Learning Model for Effective Crop Pest and Disease Detection. *Journal of Imaging, 10*(11), 279. https://doi.org/10.3390/jimaging10110279
Zhang, Z., Ma, L., Wei, C., Yang, M., Qin, S., Lv, X., & Zhang, Z. (2023). Cotton Fusarium wilt diagnosis based on generative adversarial networks in small samples. *Frontiers in Plant Science, 14*, 1290774. https://doi.org/10.3389/fpls.2023.1290774
Zhao, J., Mathieu, M., Goroshin, R., & LeCun, Y. (2016). Stacked What-

Where Auto-encoders. *arXiv*, arXiv:1506.02351. https://doi.org/10.48550/arXiv.1506.02351

Zhao, Y., Chen, Z., Gao, X., Song, W., Xiong, Q., & Hu, J. (2022). Plant Disease Detection Using Generated Leaves Based on DoubleGAN. *IEEE/ACM Transactions on Computational Biology and Bioinformatics, 19*(3), 1817-1826. https://doi.org/10.1109/TCBB.2021.3056683

Zhao, Y., Jiang, C., Wang, D., Liu, X., Song, W., & Hu, J. (2023). Identification of Plant Disease Based on Multi-Task Continual Learning. *Agronomy, 13*(12), 2863. https://doi.org/10.3390/agronomy13122863

Zheng, C., Wu, G., & Li, C. (2023). Toward Understanding Generative Data Augmentation. *arXiv*, arXiv:2305.17476. https://doi.org/10.48550/arXiv.2305.17476

Zhu, J.-Y., Park, T., Isola, P., & Efros, A. A. (2017). Unpaired Image-to-Image Translation Using Cycle-Consistent Adversarial Networks. *2017 IEEE International Conference on Computer Vision (ICCV)*, 2242-2251. https://doi.org/10.1109/ICCV.2017.244

Zilvan, V., Ramdan, A., Suryawati, E., Kusumo, R. B. S., Krisnandi, D., & Pardede, H. F. (2019). Denoising Convolutional Variational Autoencoders-Based Feature Learning for Automatic Detection of Plant Diseases. *2019 3rd International Conference on Informatics and Computational Sciences (ICICoS)*, 1-6. https://doi.org/10.1109/ICICoS48119.2019.8982494